\documentclass[10pt,twocolumn,letterpaper]{article}

\usepackage[pagenumbers]{cvpr} % To force page numbers, e.g. for an arXiv version

\usepackage[dvipsnames]{xcolor}

\usepackage{graphicx}
\usepackage{amsmath,amssymb}
\usepackage{multirow}
\usepackage{marvosym}
\usepackage{makecell}
\usepackage{bm}
\usepackage{float}

\newcommand{\PreserveBackslash}[1]{\let\temp=\\#1\let\\=\temp}
\newcolumntype{C}[1]{>{\PreserveBackslash\centering}p{#1}}
\newcolumntype{R}[1]{>{\PreserveBackslash\raggedleft}p{#1}}
\newcolumntype{L}[1]{>{\PreserveBackslash\raggedright}p{#1}}
\definecolor{cvprblue}{rgb}{0.21,0.49,0.74}
\usepackage[pagebackref,breaklinks,colorlinks,citecolor=cvprblue]{hyperref}
\newcommand*\samethanks[1][\value{footnote}]{\footnotemark[#1]}

\title{Adapting Visual-Language Models for \\Generalizable Anomaly Detection in Medical Images}

\author{Chaoqin Huang\textsuperscript{1,2,3}\thanks{Equal Contribution},
Aofan Jiang\textsuperscript{1,3}$^*$,
Jinghao Feng\textsuperscript{1,3},
Ya Zhang\textsuperscript{1,3},
Xinchao Wang\textsuperscript{2,\thanks{Corresponding authors: Yanfeng Wang and Xinchao Wang}},
Yanfeng Wang\textsuperscript{1,3,\samethanks}\\
  \textsuperscript{1} Shanghai Jiao Tong University, \textsuperscript{2} National University of Singapore \\ \textsuperscript{3} Shanghai Artificial Intelligence Laboratory\\
\tt\scriptsize \{huangchaoqin,stillunnamed,fjh1345528968,ya\_zhang,wangyanfeng622\}@sjtu.edu.cn; 
\{xinchao\}@nus.edu.sg
}

\begin{document}
\maketitle
\begin{abstract}
Recent advancements in large-scale visual-language pre-trained models have led to significant progress in zero-/few-shot anomaly detection within natural image domains. However, the substantial domain divergence between natural and medical images limits the effectiveness of these methodologies in medical anomaly detection. This paper introduces a novel lightweight multi-level adaptation and comparison framework to repurpose the CLIP model for medical anomaly detection. Our approach integrates multiple residual adapters into the pre-trained visual encoder, enabling a stepwise enhancement of visual features across different levels. This multi-level adaptation is guided by multi-level, pixel-wise visual-language feature alignment loss functions, which recalibrate the model’s focus from object semantics in natural imagery to anomaly identification in medical images. The adapted features exhibit improved generalization across various medical data types, even in zero-shot scenarios where the model encounters unseen medical modalities and anatomical regions during training. Our experiments on medical anomaly detection benchmarks demonstrate that our method significantly surpasses current state-of-the-art models, with an average AUC improvement of 6.24\% and 7.33\% for anomaly classification, 2.03\% and 2.37\% for anomaly segmentation, under the zero-shot and few-shot settings, respectively. Source code is available at: \url{https://github.com/MediaBrain-SJTU/MVFA-AD}
\end{abstract}    
\section{Introduction}
\label{sec:intro}

Medical anomaly detection (AD), which focuses on identifying unusual patterns in medical data, is central to preventing misdiagnoses and facilitating early interventions~\cite{fernando2021deep,zhang2020viral,su2021few}. The vast variability in medical images, both in terms of modalities and anatomical regions, necessitates a model that is versatile across various data types. The few-shot AD approaches~\cite{DRA, BGAD,TDG,regad} strive to attain model generalization with scarce training data, embodying a preliminary attempt for a universal AD model, despite the need for lightweight re-training~\cite{DRA,BGAD,TDG} or distribution adjustment~\cite{regad} for each new AD task.

\begin{figure}[t]
    \centering    
    \includegraphics[width=0.47\textwidth]{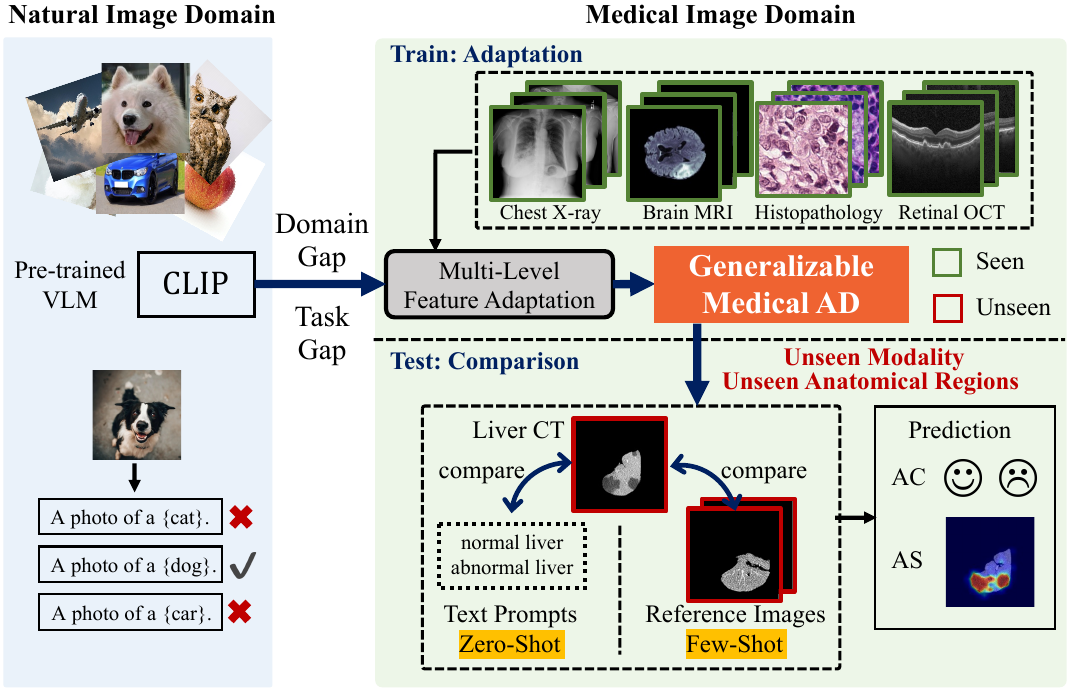}
    \vspace{-2pt}
    \caption{The overview of adaptation in pre-trained visual-language models for zero-/few-shot medical anomaly classification (AC) and anomaly segmentation (AS).}
    \label{fig:intro}
    \vspace{-15pt}
\end{figure}

Contemporary large-scale pre-trained visual-language models (VLMs) have recently paved the way for robust and generalizable anomaly detection. A notable initial effort is to directly adopt CLIP~\cite{CLIP}, a representative open-source VLM for natural imagery, for AD, simply by carefully crafting artificial text prompts~\cite{winclip}. By further employing annotated training data, Chen \emph{et al.}~\cite{chen2023zero} introduces extra linear layers to map the image features to the joint embedding space to the text features, facilitating their comparison. Despite the promise of the above two approaches, their extension to the medical domain has not been explored.

This paper attempts to develop a universal generalizable AD model for medical images, designed to be adaptable to previously unseen modalities and anatomical regions. The creation of such a model holds significant practical importance, but tailoring the CLIP model for this purpose presents a triad of challenges. Firstly, re-purposing CLIP for AD signifies a substantial shift in task requirements. The visual encoder in CLIP is known to primarily capture image semantics, yet a universal AD model must discern irregularities across diverse semantic contexts. Secondly, the transition from using CLIP in the realm of natural imagery to medical imagery constitutes a significant domain shift. Finally, the task of extending the AD model's applicability to unencountered imaging modalities and anatomical regions during the training phase is notably demanding.
 
This paper proposes a lightweight \textit{Multi-level Adaptation and Comparison} framework to re-purpose CLIP for AD in medical images as shown in Figure~\ref{fig:intro}. A multi-level visual feature \textit{adaptation} architecture is designed to align CLIP's features to the requirements of AD in medical contexts. The process of visual feature adaptation merges adapter tuning with multi-level considerations. This is achieved by integrating multiple residual adapters into the pre-trained visual encoder. This stepwise enhancement of visual features across different levels is guided by multi-level, pixel-wise visual-language feature alignment loss functions. These adapters recalibrate the model's focus, shifting it from object semantics to identifying anomalies in images, utilizing text prompts that broadly categorize images as `normal' or `anomalous'. During testing, \textit{comparison} is performed between the adapted visual features and text prompt features, and additional referenced image features if available, enabling the generation of multi-level anomaly score maps. 

The methods are evaluated on a challenging medical AD benchmark, encompassing datasets from five distinct medical modalities and anatomical regions: brain MRI ~\cite{baid2021rsna,bakas2017advancing,menze2014multimodal}, liver CT~\cite{landman2015miccai,bilic2023liver}, retinal OCT~\cite{hu2019automated,kermany2018identifying}, chest X-ray~\cite{wang2017chestx}, and digital histopathology~\cite{bejnordi2017diagnostic}. Our method outperforms state-of-the-art approaches, showcasing an average improvement of 6.24\% and 7.33\% in anomaly classification, and 2.03\% and 2.37\% in anomaly segmentation under zero-shot and few-shot scenarios, respectively. 

The main contributions are summarized below:
\begin{itemize}
    \item 
    A novel multi-level feature adaptation framework is proposed, which is, to the best of our knowledge, the first attempt to adapt pre-trained visual-language models for medical AD in zero-/few-shot scenarios. 
    \item Extensive experiments on a challenging benchmark for AD in medical images have demonstrated its exceptional generalizability across diverse data modalities and anatomical regions.
\end{itemize}
\section{Related Works}
\label{sec:related}

\noindent\textbf{Vanilla Anomaly Detection.}
Given the limited availability and high cost of abnormal images, a portion of current research on AD focuses on unsupervised methods relying exclusively on normal images~\cite{Sabokrou2018Adversarially,gong2019memorizing,US,metaformer,cutpaste,GP,ARNet,gudovskiy2022cflow,deng2022anomaly,patchcore,MKD}. Approaches such as PatchCore~\cite{patchcore} create a memory bank of normal embeddings and detect anomalies based on the distance from a test sample to the nearest normal embedding. Another method, CflowAD~\cite{gudovskiy2022cflow}, projects normal samples onto a Gaussian distribution using normalizing flows. However, relying solely on normal samples can result in an ambiguous decision boundary and reduced discriminability~\cite{bergmann2019mvtec}. In practical scenarios, a small number of anomaly samples are usually available, and these can be used to enhance detection effectiveness.

\noindent\textbf{Zero-/Few-Shot Anomaly Detection.} The utilization of a few known anomalies during training can present challenges, potentially biasing the model and hindering generalization to unseen anomalies. DRA~\cite{DRA} and BGAD~\cite{BGAD} introduce methods to mitigate this issue. Beyond simply maximizing the separation of abnormal features from normal patterns~\cite{SAD,DEVNET}, DRA~\cite{DRA} learns disentangled representations of anomalies to enable generalizable detection, accounting for unseen anomalies. BGAD~\cite{BGAD} proposes a boundary-guided semi-push-pull contrastive learning mechanism to further alleviate the bias issue. Recent advancements like WinCLIP~\cite{winclip} explore the use of foundation models for zero-/few-shot AD, leveraging language to assist in AD. Building upon WinCLIP, April-GAN~\cite{chen2023zero} maps visual features extracted from CLIP to the linear space where the text features are located, supervised by pixel-level annotated data. This paper concentrates on medical AD, a more challenging area than traditional industrial AD due to the larger gap between different data modalities.

\noindent\textbf{Medical Anomaly Detection.}
Current medical AD methods typically treat AD as a one-class classification issue, relying on normal images for training~\cite{zhou2020encoding,zhang2020viral,zhou2021proxy,zhou2021memorizing,bao2023bmad,cai2023dual,jiang2023multi}. These methods, which identify anomalies as deviations from the normal distribution, often require a large number of normal samples per class, making them impractical for real-world diagnosis. Many of these techniques are designed for a particular anatomical region~\cite{ding2022unsupervised,xu2022afsc} or are restricted to handling one specific data type per model~\cite{wolleb2022diffusion,huang2022lesionpaste,li2021unsupervised}. These methods often fall short in terms of generalizing across diverse data modalities and anatomical regions~\cite{zhang2023grace}, a pivotal aspect our paper aims to address.

\begin{figure*}[t]
    \centering    \includegraphics[width=1.0\textwidth]{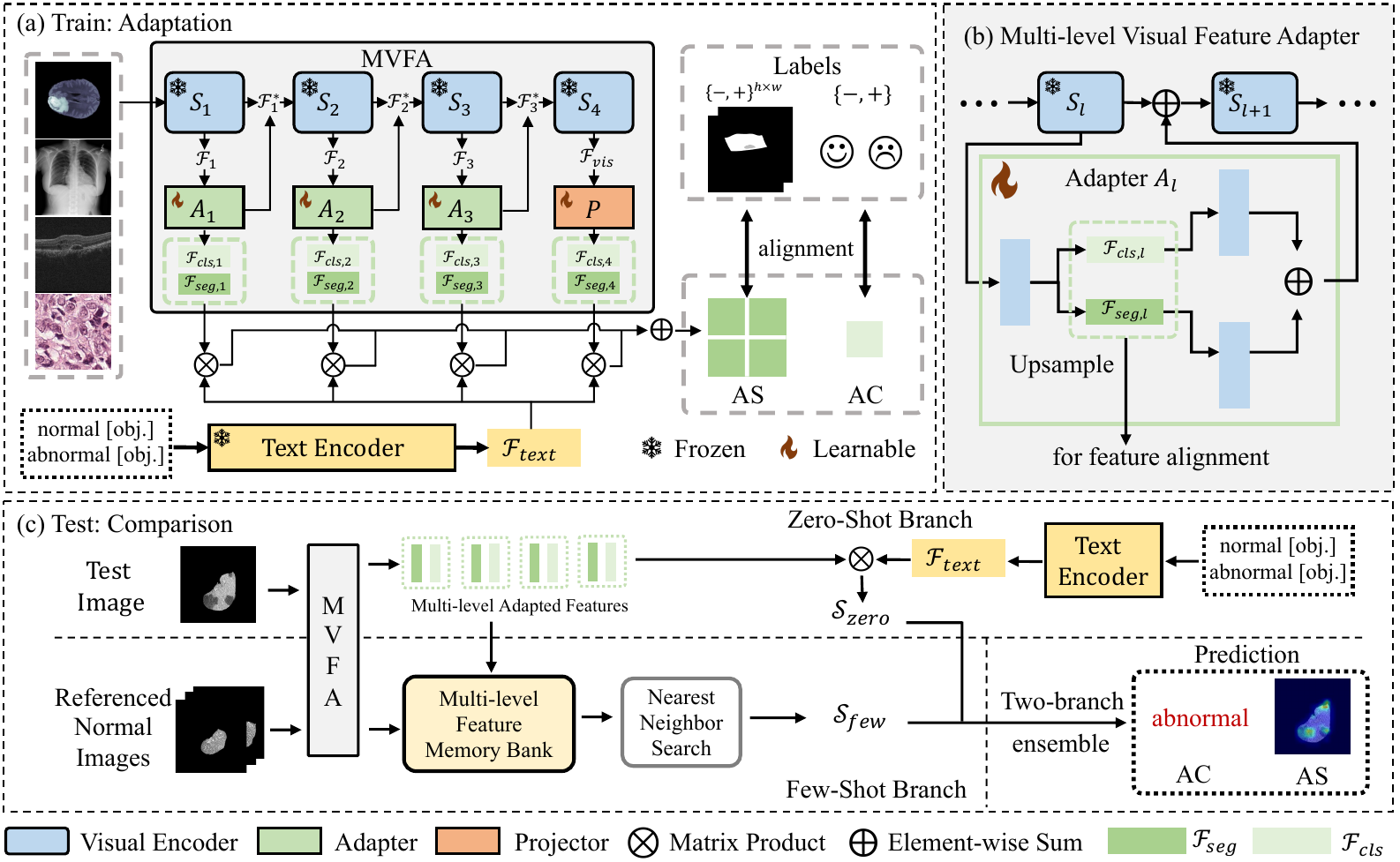}
    \caption{The architecture of multi-level adaptation and comparison framework for zero-/few-shot medical anomaly detection.}
    \vspace{-7pt}
    \label{fig:pipeline}
\end{figure*}

\noindent\textbf{Visual-Language Modeling.} 
Recently, VLMs have witnessed substantial advancements, applied to many different scenarios~\cite{YuTaoCVPR23,LiXinNeurIPS23,XingyiICCV23,JingwenAAAI24,DeepCacheCVPR24,KejiNeurIPS23}. Trained on a vast amount of image-text data, CLIP~\cite{CLIP} excels in generalizability and robustness, notably enabling language-driven zero-shot inference~\cite{taori2020measuring,goh2021multimodal}. To broaden the application of VLMs, resources like the extensive LAION-5B dataset~\cite{schuhmann2022laion} and the OpenCLIP codebase~\cite{openclip} have been made available openly. Subsequent research has underscored CLIP's potential for zero-/few-shot transfer to downstream tasks beyond mere classification~\cite{gu2021open,rombach2022high,zhou2022learning}. Several studies~\cite{rao2022denseclip,zhong2022regionclip,huang2023multi} have leveraged pre-trained CLIP models for language-guided detection and segmentation, achieving promising outcomes. This research extends the application of VLMs, initially trained on natural images, to AD in medical images, introducing a unique approach of multi-level visual feature adaptation and comparison framework.
\section{Problem Formulation}

We aim to adapt a visual-language model, initially trained on natural images (denoted as $\mathcal{M}_{nat}$), for anomaly detection (AD) in medical images, resulting in a medically adapted model ($\mathcal{M}_{med}$). This adaptation utilizes a medical training dataset $\mathcal{D}_{med}$, which consists of annotated samples from the medical field, enabling the transformation of $\mathcal{M}_{nat}$ into $\mathcal{M}_{med}$. Specifically, $\mathcal{D}_{med}$ is defined as a set of tuples $\{(x_i,c_i,\mathcal{S}_i)\}_{i=1}^{K}$, where $K$ is the total number of image samples in the dataset. Each tuple includes a training image $x_i$, its corresponding image-level anomaly classification (AC) label $c_i \in \{-, +\}$, and pixel-level anomaly segmentation (AS) annotations $\mathcal{S}_i\in \{-, +\}^{h\times w}$ for images of size $h\times w$. The label ‘$+$’ indicates an anomalous sample, while ‘$-$’ denotes a normal one. For a given test image $x_{\text{test}}$, the model aims to accurately predict both image-level and pixel-level anomalies for AC and AS, respectively.

To model the detection of anomalies from unseen imaging modalities and anatomical regions, we approach the problem in a zero-shot learning context. Here, $\mathcal{D}_{med}$ is a pre-training dataset that is composed of medical data from different modalities and anatomical regions than those in the test samples, which assesses the model's generalization to unseen scenarios. Considering the practicality of obtaining a limited number of samples from the target scenario, we also extend the method to the few-shot learning context. Here, $\mathcal{D}_{med}$ includes a small collection of $K$ annotated images that are of the same modality and anatomical region as those in the test samples, with $K$ typically representing a small numerical value, such as $\{2,4,8,16\}$ in this study.

Below we introduce our proposed multi-level adaptation and comparison framework for AD in medical images, comprising (i) multi-level feature adaptation (Sec.~\ref{sec:train}), and (ii) multi-level feature comparison (Sec.~\ref{sec:test}).

\section{Train: Multi-Level Feature Adaptation}
\label{sec:train}
To adapt a pre-trained natural image visual-language model for anomaly detection (AD) in medical imaging, we introduce a multi-level feature adaptation framework specifically designed for AD in medical images, utilizing minimal data and lightweight multi-level feature adapters.

\noindent\textbf{Multi-level Visual Feature Adapter (MVFA).}
Addressing the challenge of overfitting due to a high parameter count and limited training examples, we apply the CLIP adapter across multiple feature levels. This method appends a small set of learnable bottleneck linear layers to the visual branches of CLIP while keeping its original backbone unchanged, thus enabling adaptation at multiple feature levels. 
 
As shown in Figure~\ref{fig:pipeline} (a), for an image $x \in \mathbb{R}^{h\times w\times 3}$, a CLIP visual encoder with four sequential stages ($S_1$ to $S_4$) transforms the image $x$ into a feature space $\mathcal{F}_{vis} \in \mathbb{R}^{G\times d}$. Here, $G$ represents the grid number, and $d$ signifies the feature dimension. The output of the first three visual encoder stages ($S_1$ to $S_3$), denoted as $\mathcal{F}_l \in \mathbb{R}^{G\times d}, l\in\{1,2,3\}$, represents the three middle-stage features.

The visual feature adaptation involves three feature adapters, $A_l(\cdot)$, $l\in\{1,2,3\}$, and one feature projector, $P(\cdot)$, at different levels. At each level $l\in\{1,2,3\}$, a learnable feature adapter $A_l(\cdot)$ is integrated into the feature $\mathcal{F}_l$, encompassing two (the minimum number) layers of linear transformations. This integration transforms the features for adaptation, represented as:
\begin{equation}\label{eq:ada}
    A_l(\mathcal{F}_l) = ReLU(\mathcal{F}_l^T W_{l,1})W_{l,2}, \text{where}~l\in\{1,2,3\}.
\end{equation}
Here, $W_{l,1}$ and $W_{l,2}$ denote the learnable parameters of the linear transformations. Consistent with~\cite{gao2023clip}, a residual connection is employed in the feature adapter to retain the original knowledge encoded by the pre-trained CLIP. Specifically, a constant value $\gamma$ serves as the residual ratio to adjust the degree of preserving the original knowledge for improved performance. Therefore, the feature adapter at the $l$-th feature level can be expressed as:
\begin{equation}
    \mathcal{F}_l^* =\gamma A_l(\mathcal{F}_l)^T + (1-\gamma)\mathcal{F}_l,\text{where}~l\in\{1,2,3\},
\end{equation}
with $\mathcal{F}_l^*$ serving as the input for the next encoder stage $S_{l+1}$. By default, we set $\gamma=0.1$. Moreover, as shown in Figure~\ref{fig:pipeline} (b), to simultaneously address both global and local features for AC and AS respectively, a dual-adapter architecture replaces the single-adapter used in Eq.~\eqref{eq:ada}, producing two parallel sets of features at each level, $\mathcal{F}_{cls,l}$ and $\mathcal{F}_{seg,l}$. For the final visual feature $\mathcal{F}_{vis}$ generated by the CLIP visual encoder, a feature projector $P(\cdot)$ projects it using linear layers with parameters $W_{cls}$ and $W_{seg}$, obtaining global and local features as $\mathcal{F}_{cls,4}=\mathcal{F}_{vis}^T W_{cls}$ and $\mathcal{F}_{seg,4}=\mathcal{F}_{vis}^T W_{seg}$. Utilizing the multi-level adapted features, the model is equipped to effectively discern both global anomalies for classification and local anomalies for segmentation, through the following visual-language feature alignment.

\noindent\textbf{Language Feature Formatting.} 
To develop an effective framework for anomaly classification and segmentation, we adopt a two-tiered approach for text prompts, inspired by methodologies used in~\cite{winclip,chen2023zero}. These methods leverage descriptions of both normal and abnormal objects. At the state level, our strategy involves using straightforward, generic text descriptions for normal and abnormal states, focusing on clarity and avoiding complex details. Moving to the template level, we conduct a thorough examination of the 35 templates referenced in~\cite{deng2009imagenet} (detailed in Appendix~\ref{sec:ap_text}). By calculating the average of the text features extracted by the text encoder for normal and abnormal states separately, we obtain a text feature represented as $\mathcal{F}_{text} \in \mathbb{R}^{2\times d}$, where $d$ is the feature dimension. 

\begin{table*}[t]
\centering
\caption{Comparisons with state-of-the-art \textbf{few-shot} anomaly detection methods with K=4. The AUCs (in \%) for anomaly classification (AC) and anomaly segmentation (AS) are reported. The best result is in bold, and the second-best result is underlined.}
\label{tal:few}
\small
\setlength{\tabcolsep}{1.7pt}{
\begin{tabular}{C{2.5cm}C{2.2cm}C{1.8cm}|C{1.1cm}C{1.2cm}C{1.2cm}|C{1.0cm}C{1.0cm}C{1.0cm}C{1.0cm}C{1.0cm}C{1.0cm}}
\toprule
\multirow{2}{*}{Setting} & \multirow{2}{*}{Method} & \multirow{2}{*}{Source} & HIS & ChestXray & OCT17 & \multicolumn{2}{c}{BrainMRI} & \multicolumn{2}{c}{LiverCT} & \multicolumn{2}{c}{RESC}\\
\cmidrule(lr){4-12} 
& & & AC & AC & AC & AC & AS & AC & AS & AC & AS\\
\cmidrule(lr){1-12} 
\multirow{4}{*}{full-normal-shot} 
& CFlowAD~\cite{gudovskiy2022cflow} & WACV 2022 & 54.54 & 71.44 & 85.43 & 73.97 & 93.52 & 49.93 & 92.78 & 74.43 & 93.75 \\
& RD4AD~\cite{deng2022anomaly} & CVPR 2022 & 66.59 & 67.53 & 97.24 & 89.38 & 96.54 & 60.02 & 95.86 & 87.53 & 96.17\\
& PatchCore~\cite{patchcore} & CVPR 2022 & 69.34 & 75.17 & 98.56 & \underline{91.55} & \underline{96.97} & 60.40 & 96.58 & 91.50 & 96.39\\
& MKD~\cite{MKD} & CVPR 2022 & \underline{77.74} & \underline{81.99} & 96.62 & 81.38 & 89.54 & 60.39 & 96.14 & 88.97 & 86.60 \\
\cmidrule(lr){1-12}
\multirow{2}{*}{few-normal-shot} 
& CLIP~\cite{openclip} & OpenCLIP & 63.48 & 70.74 & 98.59 & 74.31 & 93.44 & 56.74 & 97.20 & 84.54 & 95.03\\
& MedCLIP~\cite{wang2022medclip} & EMNLP 2022 & 75.89 & \textbf{84.06} & 81.39 & 76.87	& 90.91 & 60.65 & 94.45 & 66.58 & 88.98\\
& WinCLIP~\cite{winclip} & CVPR 2023 & 67.49 & 70.00 & 97.89 & 66.85 & 94.16 & 67.19 & 96.75 & 88.83 & 96.68\\
\cmidrule(lr){1-12}
\multirow{4}{*}{\makecell[c]{few-shot}}
& DRA~\cite{DRA} & CVPR 2022 & 68.73 & 75.81 & 99.06 & 80.62 & 74.77 & 59.64 & 71.79 & 90.90 & 77.28\\
& BGAD~\cite{BGAD} & CVPR 2023 & - & - & - & 83.56 & 92.68 & \underline{72.48} & \underline{98.88} & 86.22 & 93.84\\
& APRIL-GAN~\cite{chen2023zero} & arXiv 2023 & 76.11 & 77.43 & \textbf{99.41} & 89.18 & 94.67 & 53.05 & 96.24 & \underline{94.70} & \underline{97.98} \\
\cmidrule(lr){2-12}
& MVFA & Ours & \textbf{82.71} & 81.95 & \underline{99.38} & \textbf{92.44} & \textbf{97.30} & \textbf{81.18} & \textbf{99.73} & \textbf{96.18} & \textbf{98.97}\\
\bottomrule
\end{tabular}
}
\end{table*}

\noindent\textbf{Visual-Language Feature Alignment.} 
For the image-level anomaly annotation $c \in \{-,+\}$ and the corresponding pixel-level anomaly map $\mathcal{S} \in \{-,+\}^{h\times w}$, we optimize the model at each feature level, $l \in \{1,2,3,4\}$, by aligning the adapted-visual features given by MVFA and the text features. This is achieved through a loss function that combines different components: 
\begin{equation}
\begin{aligned}
    \mathcal{L}_l = &\lambda_1\textit{Dice}(\textit{softmax}(\mathcal{F}_{seg,l} \mathcal{F}_{text}^{T}), \mathcal{S})+ \\
    &\lambda_2\textit{Focal}(\textit{softmax}(\mathcal{F}_{seg,l} \mathcal{F}_{text}^{T}), \mathcal{S})+\\
    &\lambda_3\textit{BCE}(\max_{h\times w}(\textit{softmax}(\mathcal{F}_{cls,l} \mathcal{F}_{text}^{T})),c),\\
\end{aligned}
\end{equation}
where $\textit{Dice}(\cdot,\cdot)$, $\textit{Focal}(\cdot,\cdot)$, and $\textit{BCE}(\cdot,\cdot)$ are dice loss~\cite{milletari2016v}, focal loss~\cite{lin2017focal}, and binary cross-entropy loss, respectively. $\lambda_1$, $\lambda_2$ and $\lambda_3$ are the individual loss weights where we set $\lambda_1=\lambda_2=\lambda_3=1.0$ as default. The overall adaptation loss $\mathcal{L}_{adapt}$ is then calculated as the sum of losses at each feature level, expressed as $\mathcal{L}_{adapt} = \sum_{l=1}^4 \mathcal{L}_l$. 

\noindent\textbf{Discussion.} 
WinCLIP~\cite{winclip} relies on the class token from pre-trained VLMs for natural image AD, with no adaptation performed. In contrast, MVFA introduces multi-level adaptation, freezing the main backbone while adapting features at each level via adapters in line with corresponding visual-language alignments. The resulting adapted features are residually integrated into subsequent encoder blocks, modifying input features of these blocks. This unique approach enables collaborative training of adapters across different levels via gradient propagation, enhancing the overall adaptation of the backbone model. As a result, unlike APRIL-GAN~\cite{chen2023zero}, which utilizes isolated feature projections that do not adapt the main backbone, MVFA leads to robust generalization in medical AD. The difference between MVFA and feature projection in~\cite{chen2023zero} is also evaluated in Sec.~\ref{sec:abl}.

\section{Test: Multi-Level Feature Comparison}
\label{sec:test}
During testing, to accurately predict anomalies at the image level (AC) and pixel level (AS), our approach incorporates a two-branch multi-level feature comparison architecture, comprising a zero-shot branch and a few-shot branch, as illustrated in Figure~\ref{fig:pipeline} (c).

\noindent\textbf{Zero-Shot Branch.} 
A test image $x_{test}$ is processed through MVFA to produce multi-level adapted features. These features are then compared with the text feature $\mathcal{F}_{text}$. Zero-shot AC and AS results, denoted as $c_{\text{zero}}$ and $\mathcal{S}_{\text{zero}}$, are calculated using average softmax scores across the four levels, 
\begin{equation}
\begin{aligned}
c_{\text{zero}} = \frac{1}{4}{\textstyle \sum_{l=1}^4}\max_{G}(\textit{softmax}(\mathcal{F}_{cls,l} \mathcal{F}_{text}^{T})),\\ \mathcal{S}_{\text{zero}} = \frac{1}{4}{\textstyle \sum_{l=1}^4}\text{BI}(\textit{softmax}(\mathcal{F}_{seg,l} \mathcal{F}_{text}^{T})).
\end{aligned}
\end{equation}
Here, $\text{BI}(\cdot)$ reshapes the anomaly map to $\sqrt{G}\times\sqrt{G}$ and restores it to the original input image resolution using bilinear interpolation, with $G$ representing the grid number.

\noindent\textbf{Few-Shot Branch.} 
All the multi-level visual features of a few labeled normal images in $\mathcal{D}_{med}$ contribute to constructing a multi-level feature memory bank $\mathcal{G}$ facilitating the feature comparison. The few-shot AC and AS scores, denoted as $c_{\text{few}}$ and $\mathcal{S}_{\text{few}}$, are derived from the minimum distance between the test feature and the memory bank features at each level, through a nearest neighbor search process, 
\begin{equation}
\begin{aligned}
    c_{\text{few}} = \frac{1}{4}{\textstyle \sum_{l=1}^4}\max_G(\min_{m\in\mathcal{G}}\textit{Dist}(\mathcal{F}_{cls,l},m)), \\ \mathcal{S}_{\text{few}} = \frac{1}{4}{\textstyle \sum_{l=1}^4}\text{BI}(\min_{m\in\mathcal{G}}\textit{Dist}(\mathcal{F}_{seg,l},m)).     
\end{aligned}
\end{equation}
Here, $\textit{Dist}(\cdot, \cdot)$ represents the cosine distance, calculated as $1-\text{cosine}(\cdot,\cdot)$. The final predicted AC and AS results combine the outcomes from both branches:
\begin{equation}
    c_{\text{pred}} = \beta_1 c_{\text{zero}}+ \beta_2 c_{\text{few}}, \mathcal{S}_{\text{pred}} = \beta_1 \mathcal{S}_{\text{zero}}+ \beta_2 \mathcal{S}_{\text{few}}.
\end{equation} 
Here, $\beta_1$ and $\beta_2$ are weighting factors for the zero-shot and few-shot branches, respectively, set to 0.5 by default. 
\begin{figure}
    \centering
    \begin{minipage}{0.49\linewidth}
        \centering
        \includegraphics[width=1.0\linewidth]{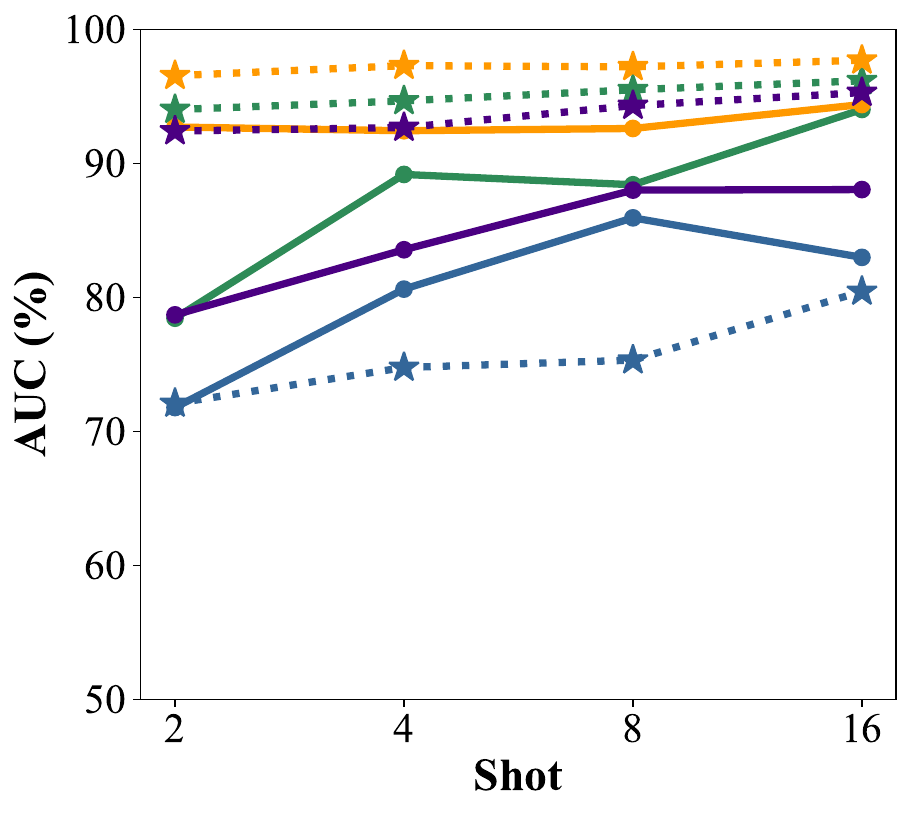}
        \subcaption*{(a) BrainMRI}
    \end{minipage}
    \begin{minipage}{0.49\linewidth}
        \centering
        \includegraphics[width=1.0\linewidth]{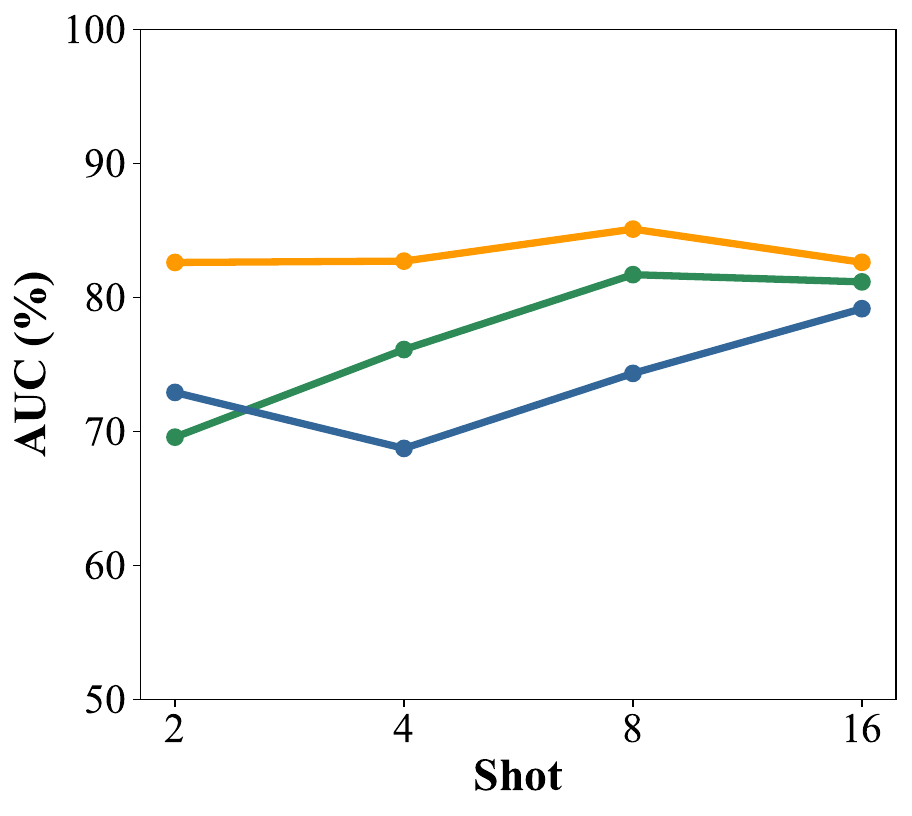}
        \subcaption*{(d) HIS}
    \end{minipage}
    \begin{minipage}{0.49\linewidth}
        \centering
        \includegraphics[width=1.0\linewidth]{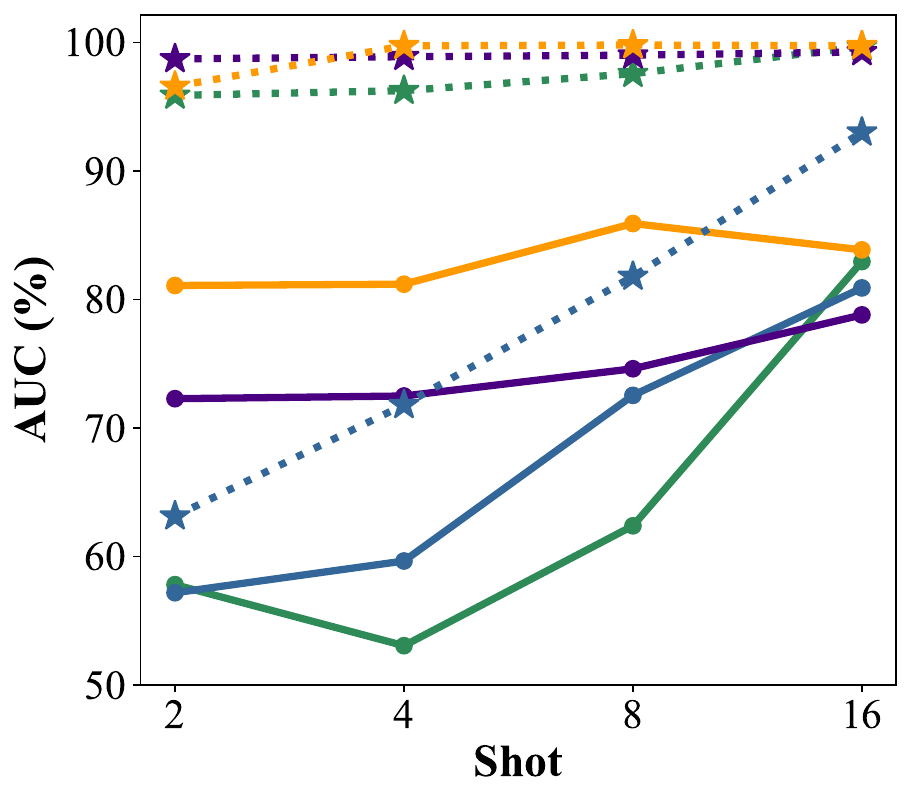}
        \subcaption*{(b) LiverCT}
    \end{minipage}
    \begin{minipage}{0.49\linewidth}
        \centering
        \includegraphics[width=1.0\linewidth]{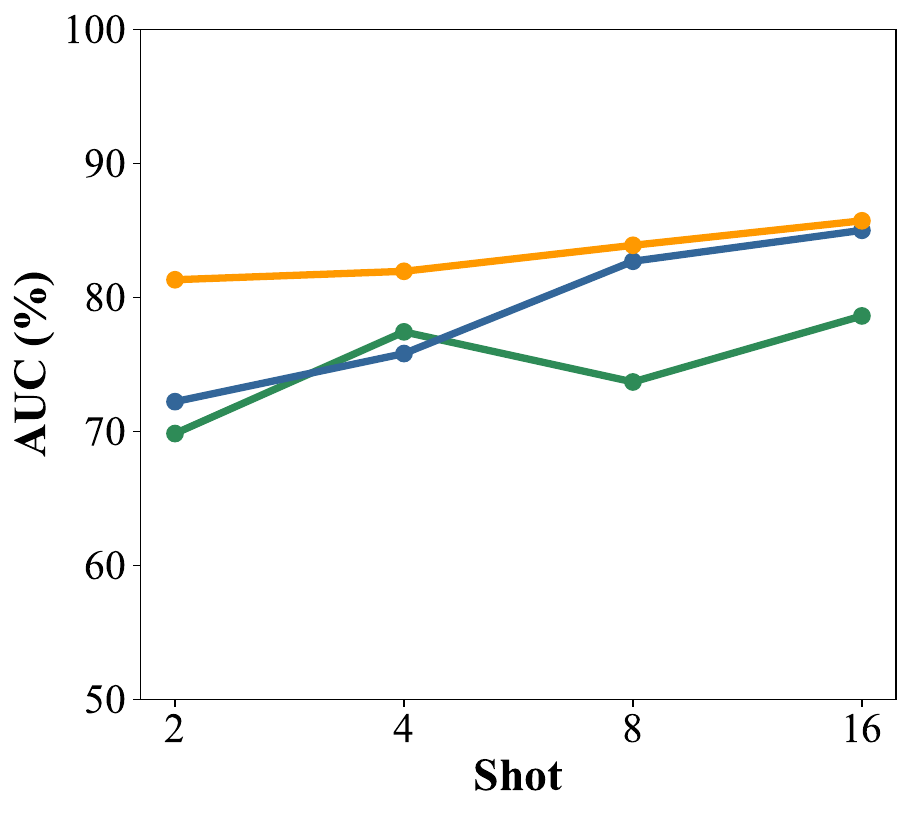}
        \subcaption*{(e) ChestXray}
    \end{minipage}
    \begin{minipage}{0.49\linewidth}
        \centering
        \includegraphics[width=1.0\linewidth]{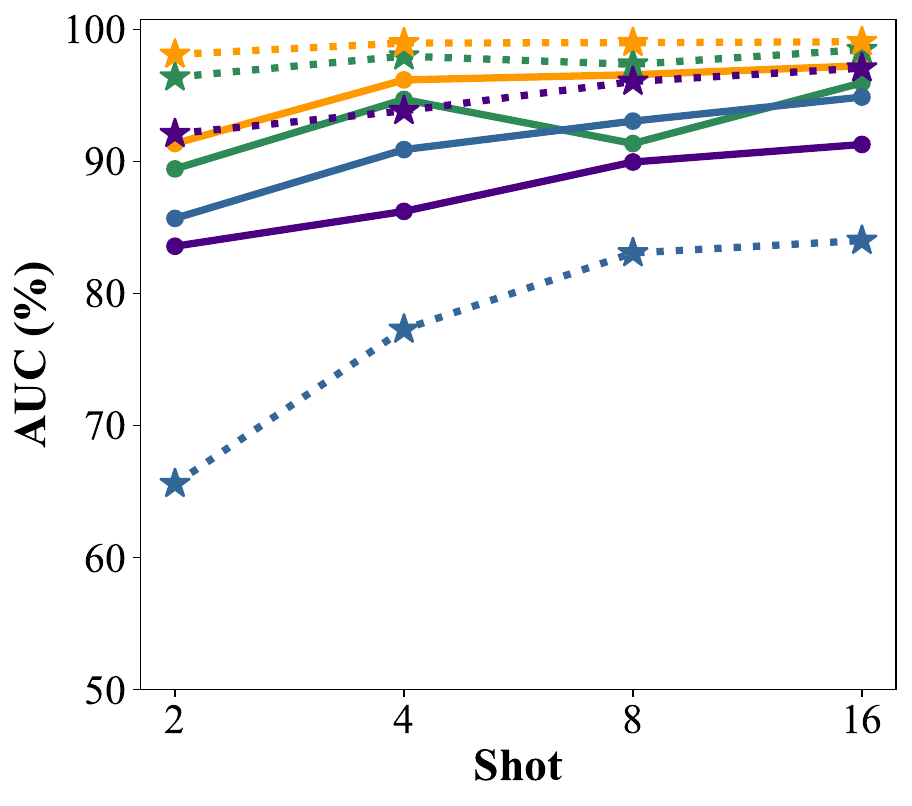}
        \subcaption*{(c) RESC}
    \end{minipage}
    \begin{minipage}{0.49\linewidth}
        \centering
        \includegraphics[width=1.0\linewidth]{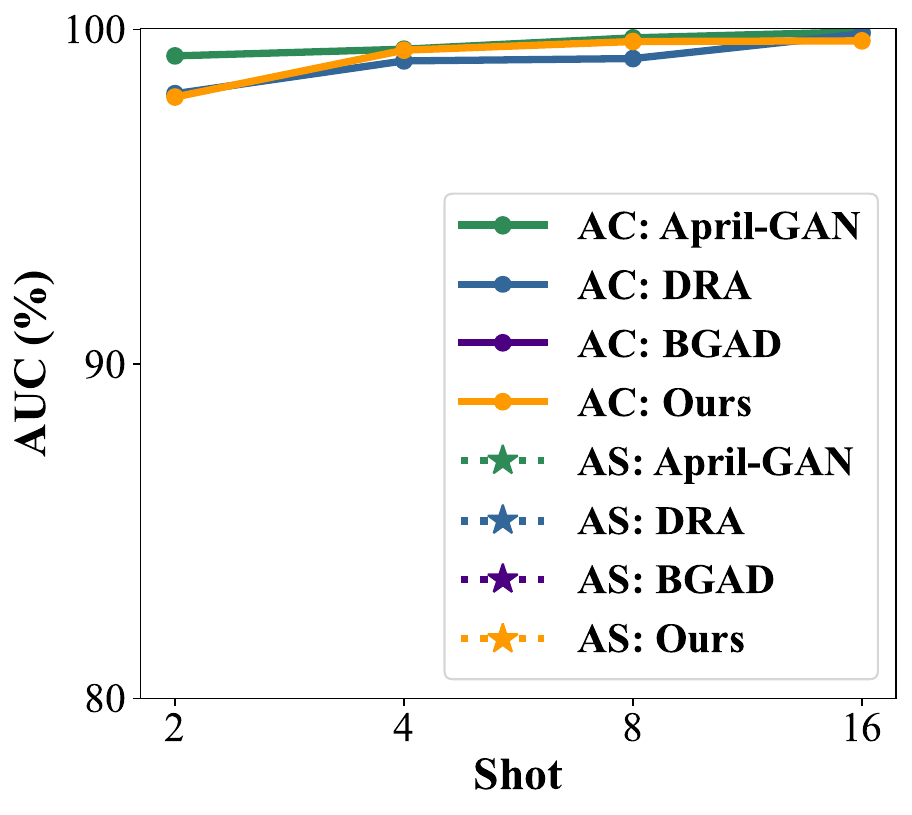}
        \subcaption*{(f) OCT17}
    \end{minipage}
    \vspace{-3pt}
    \caption{Comparisons with state-of-the-art \textbf{few-shot} anomaly detection methods on datasets of (a) BrainMRI, (b) LiverCT, (c) RESC, (d) HIS, (e) ChestXray, and (f) OCT17, with the shot number $K=\{2,4,8,16\}$. The AUCs (in \%) for anomaly classification (AC) and anomaly segmentation (AS) are reported. More details are included in Appendix~\ref{sec:ap_results}. 
    }
    \vspace{-4em}
    \label{fig:fewshot}
\end{figure}

\section{Experiments}
\label{sec:exp}

\subsection{Experimental Setups}
\noindent\textbf{Datasets.}
We consider a medical anomaly detection (AD) benchmark based on BMAD~\cite{bao2023bmad}, covering five distinct medical domains and resulting in six datasets. These include brain MRI~\cite{baid2021rsna,bakas2017advancing,menze2014multimodal}, liver CT~\cite{landman2015miccai,bilic2023liver}, retinal OCT~\cite{hu2019automated,kermany2018identifying}, chest X-ray~\cite{wang2017chestx}, and digital histopathology~\cite{bejnordi2017diagnostic}. Among these, BrainMRI~\cite{baid2021rsna,bakas2017advancing,menze2014multimodal}, LiverCT~\cite{landman2015miccai,bilic2023liver}, and RESC~\cite{hu2019automated} datasets are used for both anomaly classification (AC) and segmentation (AS), while OCT17~\cite{kermany2018identifying}, ChestXray~\cite{wang2017chestx}, and HIS~\cite{bejnordi2017diagnostic} are solely for AC. Detailed descriptions of these datasets are provided in Appendix~\ref{sec:ap_dataset}.

\noindent\textbf{Competing Methods and Baselines.}
In this study, we consider various state-of-the-art AD methods within distinct training settings as competing methods. These settings encompass (i) vanilla methods that use all normal data  (CFlowAD~\cite{gudovskiy2022cflow}, RD4AD~\cite{deng2022anomaly}, PatchCore~\cite{patchcore}, and MKD~\cite{MKD}), (ii) few-normal-shot methods (CLIP~\cite{openclip}, MedCLIP~\cite{wang2022medclip}, WinCLIP~\cite{winclip}), and (iii) few-shot methods (DRA~\cite{DRA}, BGAD~\cite{BGAD}, and April-GAN~\cite{chen2023zero}). We evaluate these methods for AC and AS, excluding BGAD, which is exclusively applied for segmentation due to its requirement for pixel-level annotations during training. 

\noindent\textbf{Evaluation Protocols.} 
The area under the Receiver Operating Characteristic curve metric (AUC) is used to quantify the performance. This metric is a standard in AD evaluation, with separate considerations for image-level AUC in AC and pixel-level AUC in AS.

\noindent\textbf{Model Configuration and Training Details.}
We utilize the CLIP with ViT-L/14 architecture, with input images at a resolution of 240. The model comprises a total of 24 layers, which are divided into 4 stages, each encompassing 6 layers. We use the Adam optimizer at a constant learning rate of 1e-3 and a batch size of 16, conducting 50 epochs for training on one single NVIDIA GeForce RTX 3090 GPU.

\subsection{Comparison with State-of-the-art Methods}
\noindent\textbf{Few-Shot Setting.} In Table~\ref{tal:few}, we compare the performance of MVFA under the few-shot setting with $K=4$ against other state-of-the-art AD methods. For an in-depth analysis of MVFA's performance across various few-shot scenarios ($K\in\{2,4,8,16\}$), please refer to Figure~\ref{fig:fewshot}. MVFA demonstrates superior performance over competing methods like DRA~\cite{DRA}, BGAD~\cite{BGAD}, and April-GAN~\cite{chen2023zero}. Notably, MVFA surpasses April-GAN, the winner of the VAND workshop at CVPR 2023~\cite{vand}, by an average of 7.33\% in AUC for AC and 2.37\% in AUC for AS across all datasets. Compared to BGAD~\cite{BGAD}, MVFA shows an average improvement of 9.18\% in AUC for AC and 3.53\% in AUC for AS, in datasets with pixel-level annotations.

MVFA outperforms few-normal-shot CLIP-based methods such as CLIP~\cite{openclip} and WinCLIP~\cite{winclip}, which also use visual-language pre-trained backbones and employ feature comparisons for AD. MVFA's advantage lies in its ability to effectively utilize a few abnormal samples, leading to significant gains over these methods. For example, against WinCLIP~\cite{winclip}, MVFA achieves an average improvement of 12.60\% in AUC for AC and 2.81\% in AUC for AS across all datasets. While MedCLIP~\cite{wang2022medclip} shows superior results on ChestXray because it was trained on large-scale overlapping ChestXray data in our medical AD benchmark, it lacks broad generalization capabilities, as evidenced by its performance on other datasets.

Moreover, MVFA exhibits substantial improvements over full-normal-shot vanilla AD methods such as CFlowAD~\cite{gudovskiy2022cflow}, RD4AD~\cite{deng2022anomaly}, PatchCore~\cite{patchcore}, and MKD~\cite{MKD}, which rely on much larger datasets than those employed in this study. This highlights the value of incorporating a few abnormal samples as supervision, especially in medical diagnostics where acquiring a limited number of abnormal data can be more practical.

\begin{table}[t]
\centering
\caption{Comparisons with state-of-the-art \textbf{zero-shot} anomaly detection methods with in-/out-domain evaluation. The AUCs (in \%) for AC and AS are reported.}
\vspace{-5pt}
\label{tal:zero}
\small
\scalebox{0.95}{
\setlength{\tabcolsep}{1.0pt}{
\begin{tabular}{C{1.5cm}|C{2.1cm}C{2.2cm}C{2.1cm}}
\toprule
Datasets & WinCLIP~\cite{winclip} & April-GAN~\cite{chen2023zero} & MVFA (ours)\\
\cmidrule(lr){1-4}
HIS & 69.85~/~- & 72.36~/~- & \textbf{77.90}~/~-\\
ChestXray & 70.86~/~- & 57.49~/~- & \textbf{71.11}~/~-\\
OCT17 & 46.64~/~- & 92.61~/~- & \textbf{95.40}~/~-\\
BrainMRI & 66.49~/~85.99 & 76.43~/~\textbf{91.79} & \textbf{78.63}~/~90.27 \\
LiverCT & 64.20~/~96.20 & 70.57~/~97.05 & \textbf{76.24}~/~\textbf{97.85} \\
RESC & 42.51~/~80.56 & 75.67~/~85.23 & \textbf{83.31}~/~\textbf{92.05} \\
\bottomrule
\end{tabular}}}
\end{table}

\begin{table}[t]
    \centering
    \caption{Comparisons with state-of-the-art \textbf{few-shot} anomaly detection methods with K=4 for in-/out-domain evaluation. The AUCs (in \%) for AC/AS are reported.}
    \vspace{-5pt}
\label{tal:in_exp}
    \small
    \scalebox{0.95}{
    \setlength{\tabcolsep}{1.5pt}{
\begin{tabular}{C{2.8cm}|C{1.8cm}C{1.9cm}C{1.7cm}}
\toprule
    AC/AS (avg AUC\%)    &  WinCLIP~\cite{winclip} & April-GAN~\cite{chen2023zero} & MVFA\\
         \cmidrule(lr){1-4}
         in-domain (MVTec) &  95.16/96.27 & 92.77/95.89 & \textbf{96.19}/\textbf{96.32}\\
       out-domain (medical) &  76.38/95.86 & 81.65/96.30 & \textbf{88.97}/\textbf{98.67}\\
       \bottomrule
    \end{tabular}}}
\end{table}

\begin{table*}[t]
\centering
\caption{Ablation studies compared with feature alignment with multi-level projectors and feature adaptation with multi-level adapters under the zero-shot setting. The AUCs (in \%) for AC and AS are reported.}
\label{tal:abl1}
\small
\setlength{\tabcolsep}{2.2pt}{
\begin{tabular}{C{4.0cm}|C{1.3cm}C{1.3cm}C{1.3cm}|C{1.3cm}C{1.3cm}C{1.3cm}C{1.3cm}C{1.3cm}C{1.3cm}}
\toprule
\multirow{2}{*}{Method} & HIS & ChestXray & OCT17 & \multicolumn{2}{c}{BrainMRI} & \multicolumn{2}{c}{LiverCT} & \multicolumn{2}{c}{RESC}\\
\cmidrule(lr){2-10} 
& AC & AC & AC & AC & AS & AC & AS & AC & AS\\
\cmidrule(lr){1-10}
feature alignment (projectors) & 66.32 & 58.06 & 49.85 & 76.66 & 89.39 & 75.85 & 97.64 & 74.44 & 89.17\\
feature adaptation (adapters) & \textbf{77.90} & \textbf{71.11} & \textbf{95.40} & \textbf{78.63} & \textbf{90.27} & \textbf{76.24} & \textbf{97.85} &  \textbf{83.31} & \textbf{92.05}\\
\bottomrule
\end{tabular}}
\end{table*}

\begin{table}[t]
\centering
\caption{Ablation studies of multi-level feature ensemble under \textit{single-level training/multi-level training} with the same model architecture. The average AUCs (in \%) of all six datasets for AC and AS under the few-shot setting (K=4) are reported. Results for each dataset are included in Appendix~\ref{sec:ap_results}.}
\label{tal:abl3}
\small
\scalebox{0.95}{
\setlength{\tabcolsep}{1.2pt}{
\begin{tabular}{C{0.6cm}|C{1.65cm}C{1.65cm}C{1.65cm}C{1.65cm}|C{1.0cm}}
\toprule
& Layer 1 & Layer 2 & Layer 3 & Layer 4 & All\\
\cmidrule(lr){1-6} 
AC & 80.96/83.39 & 88.83/88.84 & 81.25/84.32 & 83.33/84.62 & \textbf{88.97}\\
AS & 97.70/97.98 & 98.58/98.62 & 96.03/98.54 & 97.19/97.44 & \textbf{98.67}\\
\bottomrule
\end{tabular}}}
\end{table}

\noindent\textbf{Zero-Shot Setting.}
The experiments for zero-shot AD were conducted under the \textit{leave-one-out setting}. In this configuration, a designated target dataset was chosen for testing, while the remaining datasets with different modalities and anatomical regions were employed for training. The aim of this approach was to gauge the performance when confronted with unseen modalities and anatomical regions, thereby assessing the model's capacity for generalization. Table~\ref{tal:zero} provides a comprehensive overview of the results pertaining to zero-shot medical AC and AS, offering a comparative evaluation alongside two state-of-the-art methods that also harness the capabilities of the CLIP backbone. MVFA shows remarkable superiority; for instance, it outperforms WinCLIP~\cite{winclip} with an average AUC improvement of 20.34\% for AC and 5.81\% for AS across all datasets. Similarly, against April-GAN~\cite{chen2023zero}, MVFA achieves an average AUC improvement of 6.24\% for AC and 2.03\% for AS across all datasets, underscoring its effectiveness in the challenging zero-shot medical AD setting.

\noindent\textbf{In-Domain Evaluation.} The MVTec AD benchmark~\cite{bergmann2019mvtec}, consisting of 15 industrial defect detection sub-datasets, is considered as in-domain evaluation. As shown in Table~\ref{tal:in_exp}, although our main focus is not in-domain scenarios, MVFA shows comparable performance to state-of-the-art methods in a few-shot setting (K=4). Detailed results for each sub-dataset are included in Appendix~\ref{sec:ap_results}. In our main focus of out-domain evaluations on the medical AD benchmark, MVFA significantly outperforms competing methods, highlighting its superior generalization capabilities.

\subsection{Ablation Studies}\label{sec:abl}

\textbf{Feature Adaptation vs. Feature Alignment.}
We conduct ablation studies in the zero-shot setting for AC and AS to assess the effectiveness of multi-level feature adapters in enhancing cross-modal generalization. For this purpose, we substitute the multi-level feature adapters with distinct multi-level feature projectors, while keeping the parameters and feature alignment loss functions the same. The primary distinction is that each projector is optimized independently, in contrast to the collective optimization approach used for adapters. This ensures that the only variable under consideration is the architecture, while all other factors, including model parameters and training loss functions, remain constant for a valid comparison.

The results, as shown in Table~\ref{tal:abl1}, reveal significant improvements with feature adapters. They lead to substantial improvements in AC, with image-level AUC increasing by 11.58\%, 13.05\%, 45.55\%, 1.97\%, 0.39\%, and 8.87\% for HIS, ChestXay, OCT17, BrainMRI, LiverCT, and RESC, respectively, with an average improvement of 13.57\%. For AS, improvements were observed, with pixel-level AUC increasing by 0.88\%, 0.21\%, and 2.88\% for BrainMRI, LiverCT, and RESC, respectively. These findings highlight the critical role of multi-level feature adapters in boosting the model's generalization capabilities.

\begin{figure}[t]
    \centering    \includegraphics[width=0.48\textwidth]{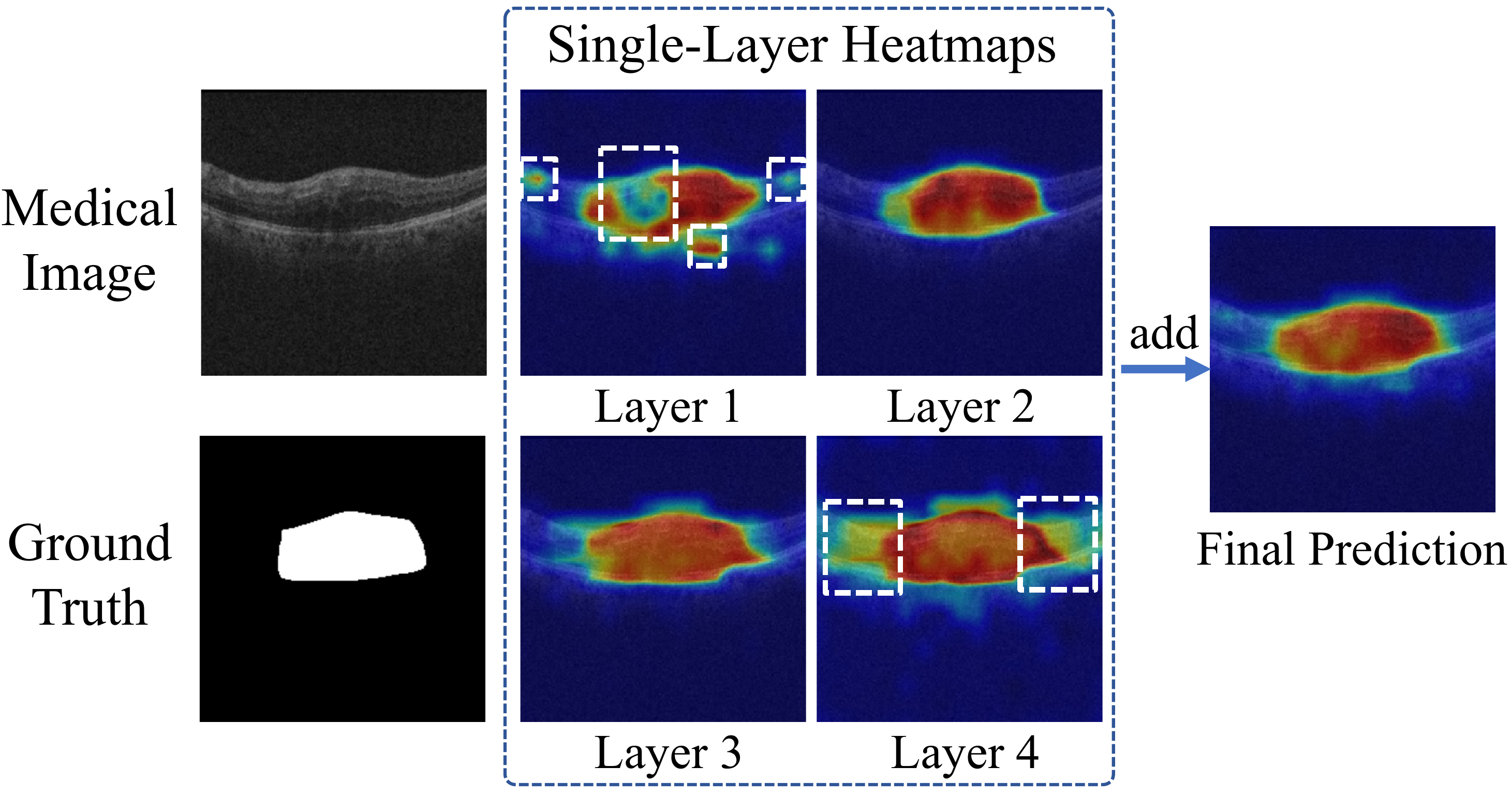}
    \caption{Considering features across multiple levels significantly enhances segmentation performance. The white dashed boxes demarcate regions that have been missed or erroneously segmented. More cases are included in Appendix~\ref{sec:ap_vis}.}
    \label{fig:intro_multilayer}
    \vspace{-5pt}
\end{figure}

\begin{figure*}[t]
    \centering
\includegraphics[width=0.9\textwidth]{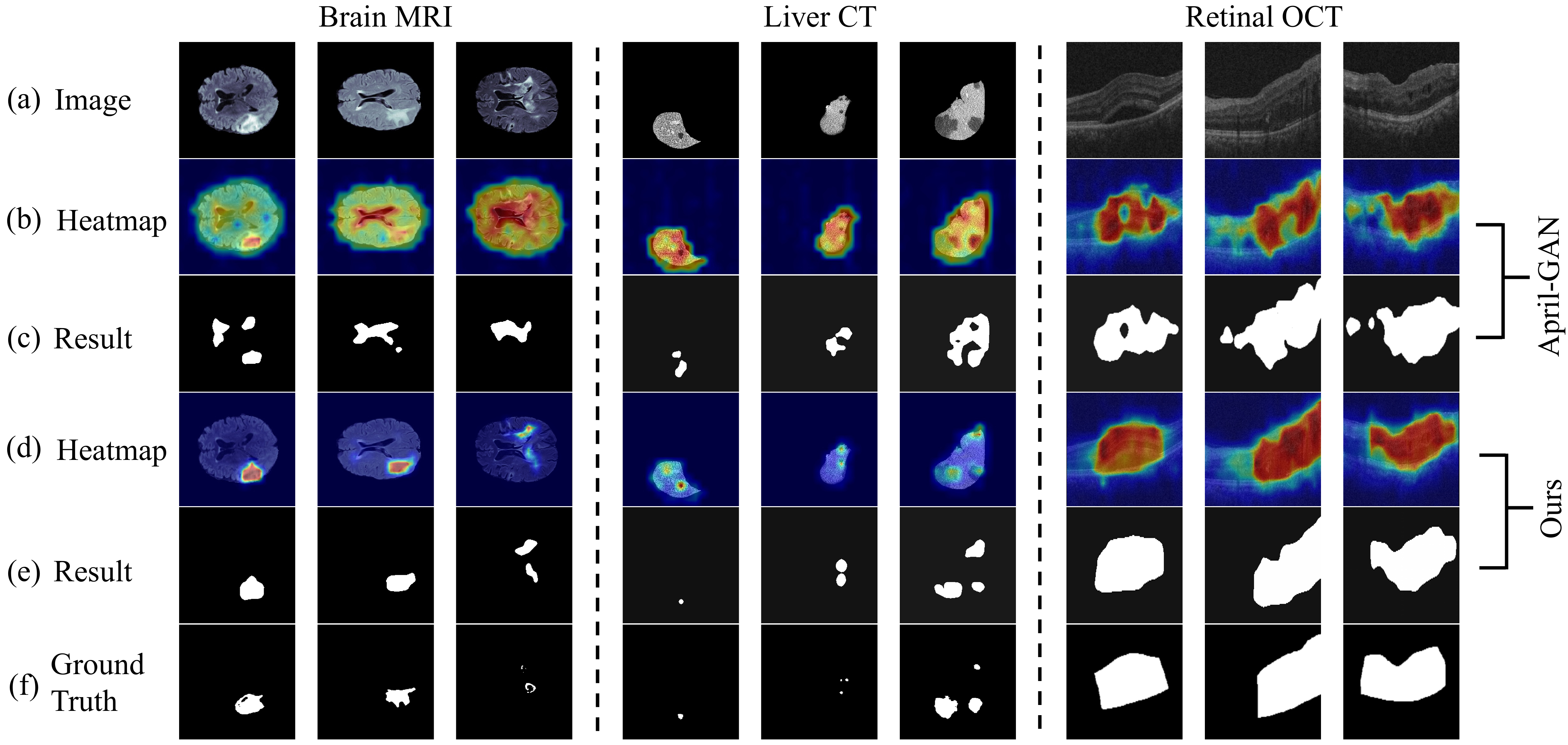}
\vspace{-4pt}
    \caption{Visualization of AS for MVFA on brain MRI, liver CT, and retinal OCT, compared with state-of-the-art method April-GAN. Results from (e) show better performance than results from (c), showing the effectiveness of the proposed multi-level feature adaptation.}
    \label{fig:vis}
    \vspace{-8pt}
\end{figure*}

\begin{figure}[t]
    \centering
    \begin{minipage}{0.49\linewidth}
        \centering
        \includegraphics[width=1.0\linewidth]{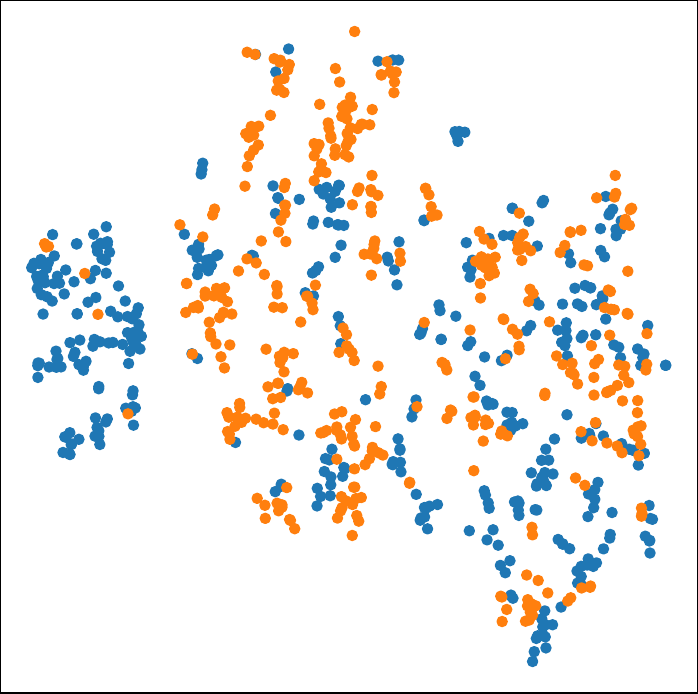}
        \subcaption*{(a) Without Adapter}
    \end{minipage}
    \begin{minipage}{0.49\linewidth}
        \centering
        \includegraphics[width=1.0\linewidth]{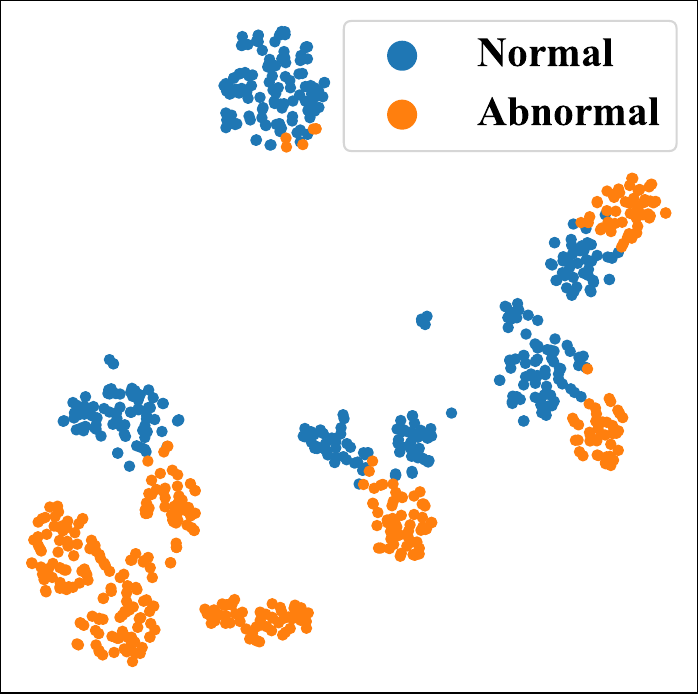}
        \subcaption*{(b) With Adapter}
    \end{minipage}
    \caption{Visualization, using t-SNE, of the features learned from the RESC test set, using (a) pretrained visual encoder, and (b) multi-level feature adapters. The same t-SNE optimization iterations are used in each case. Results show that features extracted by adapters are separated between normal and abnormal samples.}
    \vspace{-3pt}
    \label{fig:tsne}
\end{figure}

\noindent\textbf{Feature Ensemble in Multi-Level.} 
In this study, as shown in Table~\ref{tal:abl3}, we evaluate the effectiveness of an ensemble approach that integrates features from different layers. The approaches are compared against our comprehensive model, both using the multi-level adapter architecture. Our evaluation specifically focused on features from four distinct layers under two training scenarios: (i) single-layer training, where only one layer's adapter is optimized in each experiment, and (ii) multi-level training, aligning with the methodology of our comprehensive model. The objective of this comparison was to determine the benefits of combining features from multiple layers compared to optimizing each layer's features separately.

The results presented in Table~\ref{tal:abl3} demonstrate that among individual layers, Layer 2 yielded the highest performance, achieving 88.84\% in AC and 98.62\% in AS. However, the ensemble method, which integrates features from all layers, outperformed single-layer approaches, recording AUCs of 88.97\% in AC and 98.67\% in AS. This underscores the effectiveness of combining features from multiple levels. Moreover, multi-level training consistently exceeds the performance of single-layer training, reinforcing the benefits of our multi-level adaptation approach across all layers.

The visualizations of retina anomaly segmentation from various layers, as depicted in Figure~\ref{fig:intro_multilayer}, further reinforce our findings. These visualizations distinctly show that single-layer methods are less effective compared to the ensemble approaches. By synergistically integrating features from different layers, we achieve a marked improvement in AD, which aligns with and supports our quantitative results. 

\subsection{Visualization Analysis}

To qualitatively analyze how the proposed multi-level feature adaptation approach improves anomaly segmentation performance, we visualize the results of several cases from the BrainMRI, LiverCT, and RESC datasets. It can be seen from the results in Figure~\ref{fig:vis} that the segmentation produced by our MVFA (column e) is closer to the ground truth (column f) than that produced by the state-of-the-art method April-GAN (column c). This illustrates the effectiveness of the proposed multi-level visual feature adaptation.

We employ t-SNE~\cite{van2008visualizing} to visualize the learned features from RESC, as depicted in Figure~\ref{fig:tsne}. Each dot corresponds to features of a normal or abnormal sample from the test set. It can be seen that adapter enhances the separation between samples belonging to distinct states, which is beneficial for the identification of AD decision boundaries.
\section{Conclusion}
This paper adapts the pretrained visual-language models in natural domain to medical AD, with cross-domain generalizability among various modalities and anatomical regions. 
The adaptation involves not only from natural domain to medical domain, but also from high-level semantics to pixel-level segmentation. 
To achieve such goals, a collaborative multi-level feature adaptation method is introduced, where each adaptation is guided by the corresponding visual-language alignments, facilitating segmenting anomalies of diverse forms from medical images.
Coupled with a comparison-based AD strategy, the method enables flexible adaptation for datasets with substantial modality and distribution differences. 
The proposed method outperforms existing methods on zero-/few-shot AC and AS tasks, indicating promising research avenues for future exploration.

\noindent\textbf{Acknowledgment.} This work is supported by the National Key R\&D Program of China (No. 2022ZD0160703), STCSM (No. 22511106101, No. 18DZ2270700, No. 21DZ1100100), 111 plan (No. BP0719010), State Key Laboratory of UHD Video and Audio Production and Presentation,
and the Ministry of Education, Singapore, under its Academic Research Fund Tier 2 (Award Number: MOE-T2EP20122-0006).
{
\small
\bibliographystyle{ieeenat_fullname}
\bibliography{main}
}
\newpage
\begin{table*}[t]
\small
\setlength{\tabcolsep}{2.2pt}{
\begin{tabular}{C{2.0cm}C{3.0cm}|C{2.5cm}C{2.5cm}C{1.5cm}C{1.8cm}C{3.2cm}}
\toprule
     Datasets & Sources & Train (all-normal) & Train (with labels) &  Test& Sample size & Annotation Level \\
    \hline
    BrainMRI & BraTS2021~\cite{baid2021rsna,bakas2017advancing,menze2014multimodal}& 7,500 & 83 & 3,715  &240*240  & Segmentation mask   \\
    LiverCT & BTCV\cite{landman2015miccai} + LiTs~\cite{bilic2023liver} & 1,452 & 166 &1,493  & 512*512 & Segmentation mask  \\
    RESC & RESC~\cite{hu2019automated}& 4,297 & 115 & 1,805  & 512*1,024 & Segmentation mask  \\
    OCT17 & OCT2017~\cite{kermany2018identifying}& 26,315 & 32 &968 & 512*496 & Image label \\
    ChestXray & RSNA~\cite{wang2017chestx}&8,000& 1,490 &17,194 & 1,024*1,024 & Image label    \\
    HIS & Camelyon16~\cite{bejnordi2017diagnostic}&5,088&  236  &2,000 &  256*256 & Image label \\
    \bottomrule
    \end{tabular}
    }
    \caption{Summary of datasets from different medical modalities.}
\label{tal:dataset}
\end{table*}

\begin{figure*}
\noindent\begin{minipage}[t]{0.32\linewidth}
(a) \emph{State}-level (-:normal, +:abnormal)

{\tt \small
\begin{itemize}
    \item[-] c := "[o]"
    \item[-] c := "flawless [o]"
    \item[-] c := "perfect [o]"
    \item[-] c := "unblemished [o]"
    \item[-] c := "[o] without flaw"
    \item[-] c := "[o] without defect"
    \item[-] c := "[o] without damage"
    \item[+] c := "damaged [o]"
    \item[+] c := "[o] with flaw"
    \item[+] c := "[o] with defect"
    \item[+] c := "[o] with damage"
\end{itemize}
}
\end{minipage}
\hfill
\begin{minipage}[t]{0.38\linewidth}
(b) \emph{Template}-level

{\tt \small
\begin{itemize}
\item "a photo of a/the/one [c]."
\item "a photo of a/the cool [c]."
\item "a photo of a/the small [c]."
\item "a photo of a/the large [c]."
\item "a bright photo of a/the [c]." 
\item "a dark photo of a/the [c]."
\item "a blurry photo of a/the [c]."
\item "a bad photo of a/the [c]."
\item "a good photo of a/the [c]."
\item "a cropped photo of a/the [c]."
\item "a close-up photo of a/the [c]."
\end{itemize}
}
\end{minipage}
\begin{minipage}[t]{0.29\linewidth}
{\tt \small
\begin{itemize}
\item {\rm (cont'd)} "a photo of my [c]."
\item "a low resolution photo of a/the [c]."
\item "a black and white photo of a/the [c]."
\item "a jpeg corrupted photo of a/the [c]."
\item "there is a/the [c] in the scene."
\item "this is a/the/one [c] in the scene."
\end{itemize}
}
\end{minipage}
\caption{Lists of state and template level prompts employed in this paper to construct text features.}
\label{fig:comp_prompt}
\vspace{10pt}
\end{figure*}

\begin{table*}[h]
\centering
\caption{Comparisons with state-of-the-art \textbf{few-shot} anomaly detection methods with $K=2,4,8,16$. The AUCs (in \%) for anomaly classification (AC) and anomaly segmentation (AS) are reported. The best result is in bold, and the second-best result is underlined.}
\label{tal:suppfew}
\small
\setlength{\tabcolsep}{2.2pt}{
\begin{tabular}{C{2.0cm}C{2.2cm}C{1.8cm}|C{1.1cm}C{1.2cm}C{1.2cm}|C{1.0cm}C{1.0cm}C{1.0cm}C{1.0cm}C{1.0cm}C{1.0cm}}
\toprule
\multirow{2}{*}{Shot Number} & \multirow{2}{*}{Method} & \multirow{2}{*}{Source} & HIS & ChestXray & OCT17 & \multicolumn{2}{c}{BrainMRI} & \multicolumn{2}{c}{LiverCT} & \multicolumn{2}{c}{RESC}\\
\cmidrule(lr){4-12} 
& & & AC & AC & AC & AC & AS & AC & AS & AC & AS\\
\cmidrule(lr){1-12} 
& DRA~\cite{DRA} & CVPR 2022 & \underline{72.91} & \underline{72.22} & \underline{98.08} & 71.78 & 72.09 & 57.17 & 63.13 & 85.69 & 65.59 \\
& BGAD~\cite{BGAD} & CVPR 2023 & - & - & - & \underline{78.70}  & 92.42 & \underline{72.27} & \textbf{98.71} & 83.58 & 92.10 \\
& APRIL-GAN~\cite{chen2023zero} & arXiv 2023   & 69.57 & 69.84 & \textbf{99.21} & 78.45 & \underline{94.02} & 57.80 & 95.87 & \underline{89.44} & \underline{96.39} \\
\multirow{-4}{*}{2-shot} & MFA & ours & \textbf{82.61} & \textbf{81.32} & 97.98 & \textbf{92.72} & \textbf{96.55} & \textbf{81.08} & \underline{96.57} & \textbf{91.36} & \textbf{98.11} \\
\cmidrule(lr){1-12} 
& DRA~\cite{DRA} & CVPR 2022 & 68.73 & 75.81 & 99.06 & 80.62 & 74.77 & 59.64 & 71.79 & 90.90 & 77.28 \\
& BGAD~\cite{BGAD} & CVPR 2023 & - & - & - & 83.56 & 92.68 & \underline{72.48} & \underline{98.88} & 86.22 & 93.84 \\
& APRIL-GAN~\cite{chen2023zero} & arXiv 2023   & \underline{76.11} & \underline{77.43} & \textbf{99.41} & \underline{89.18} & \underline{94.67} & 53.05 & 96.24 & \underline{94.70}  & \underline{97.98} \\
\multirow{-4}{*}{4-shot} & MFA & ours & \textbf{82.71} & \textbf{81.95} & \underline{99.38} & \textbf{92.44} & \textbf{97.30}  & \textbf{81.18} & \textbf{99.73} & \textbf{96.18} & \textbf{98.97} \\
\cmidrule(lr){1-12} 
& DRA~\cite{DRA} & CVPR 2022 & 74.33 & \underline{82.70}  & 99.13 & 85.94 & 75.32 & 72.53 & 81.78 & \underline{93.06} & 83.07 \\
& BGAD~\cite{BGAD} & CVPR 2023 & - & - & - & 88.01 & 94.32 & \underline{74.60} & \underline{99.00} & 89.96 & 96.06\\
& APRIL-GAN~\cite{chen2023zero} & arXiv 2023   & \underline{81.70}  & 73.69 & \textbf{99.75} & \underline{88.41} & \underline{95.50}  & 62.38 & 97.56 & 91.36 & \underline{97.36} \\
\multirow{-4}{*}{8-shot} & MFA & ours & \textbf{85.10}  & \textbf{83.89} & \underline{99.64} & \textbf{92.61} & \textbf{97.21} & \textbf{85.90} & \textbf{99.79} & \textbf{96.57} & \textbf{99.00} \\
\cmidrule(lr){1-12} 
& DRA~\cite{DRA} & CVPR 2022 & 79.16 & \underline{85.01} & \underline{99.87} & 82.99 & 80.45 & 80.89 & 93.00 & 94.88 & 84.01 \\
& BGAD~\cite{BGAD} & CVPR 2023 & - & - & - & 88.05 & 95.29 & 78.79 & 99.25 & 91.29 & 97.07 \\
& APRIL-GAN~\cite{chen2023zero} & arXiv 2023   & \underline{81.16} & 78.62 & \textbf{99.93} & \underline{94.03} & \underline{96.17} & \underline{82.94} & \underline{99.64} & \underline{95.96} & \underline{98.47} \\
\multirow{-4}{*}{16-shot}  & MFA & ours & \textbf{82.62} & \textbf{85.72} & 99.66 & \textbf{94.40}  & \textbf{97.70}  & \textbf{83.85} & \textbf{99.73} & \textbf{97.25} & \textbf{99.07}\\
\bottomrule
\end{tabular}
}
\end{table*}

\newpage
\clearpage
\appendix
\section{Medical Anomaly Detection Benchmark}\label{sec:ap_dataset}

The details of the medical anomaly detection (AD) benchmark are concisely summarized in Table~\ref{tal:dataset}. For the few-shot AD scenario, we select a random subset of labeled training samples, with $K\in \{2,4,8,16\}$, from the labeled training set (designated as ``Train (with labels)'' in Table~\ref{tal:dataset}). These samples are employed in various competing baselines, including CLIP~\cite{openclip}, WinCLIP~\cite{winclip}, DRA~\cite{DRA}, BGAD~\cite{BGAD}, and April-GAN~\cite{chen2023zero}. Furthermore, consistent with the original methodologies that require training on a substantial amount of normal data, such as CFlowAD~\cite{gudovskiy2022cflow}, RD4AD~\cite{deng2022anomaly}, PatchCore~\cite{patchcore}, and MKD~\cite{MKD}, we employ a dataset exclusively comprising normal images. This dataset is referred to as ``Train (all-normal)'' for training purposes. It is important to highlight that this ``all-normal'' training set encompasses considerably more data compared to the limited data used in the few-shot scenario.
Below are detailed descriptions of datasets used in medical AD benchmark:

\begin{table*}[h!]
\centering
\caption{Ablation study of multi-level features \textbf{without multi-level training}. The AUCs (in \%) for classification (AC) and segmentation (AS) under the few-shot setting (k=4) are reported. The best result is in bold, and the second-best result is underlined.}
\label{tal:supplayer_2}
\small
\setlength{\tabcolsep}{2.5pt}{
\begin{tabular}{C{2.5cm}|C{1.4cm}C{1.4cm}C{1.4cm}|C{1.4cm}C{1.4cm}C{1.4cm}C{1.4cm}C{1.4cm}C{1.4cm}}
\toprule
\multirow{2}{*}{Layers} & HIS & ChestXray & OCT17 & \multicolumn{2}{c}{BrainMRI} & \multicolumn{2}{c}{LiverCT} & \multicolumn{2}{c}{RESC}\\
\cmidrule(lr){2-10} 
& AC & AC & AC & AC & AS & AC & AS & AC & AS\\
\cmidrule(lr){1-10} 
Layer 1 & 74.54 & 78.69 & 97.75 & 87.84 & 97.05 & 58.15 & 98.47 & 88.76 & 97.58\\
Layer 2 & \underline{81.36} & \underline{81.09} & \textbf{99.84} & \underline{90.81} & \textbf{97.34} & \textbf{85.36} & \underline{99.58} & \underline{94.54} & \underline{98.81}\\
Layer 3 & 69.00 & 79.75 & 98.68 & 83.01 & 94.34 & 63.78 & 95.35 & 93.29 & 98.40\\
Layer 4 & 71.02 & 72.84 & 99.37 & 86.92 & 95.42 & 76.09 & 97.92 & 93.72 & 98.23\\
\cmidrule(lr){1-10} 
Ensemble & \textbf{82.71} & \textbf{81.95} & \underline{99.38} & \textbf{92.44} & \underline{97.30} & \underline{81.18} & \textbf{99.73} & \textbf{96.18} & \textbf{98.97}\\
\bottomrule
\end{tabular}
}
\end{table*}

\begin{table*}[h!]
\centering
\caption{Ablation study of multi-level features \textbf{with multi-level training}. The AUCs (in \%) for classification (AC) and segmentation (AS) under the few-shot setting (k=4) are reported. The best result is in bold, and the second-best result is underlined.}
\label{tal:supplayer}
\small
\setlength{\tabcolsep}{2.5pt}{
\begin{tabular}{C{2.5cm}|C{1.4cm}C{1.4cm}C{1.4cm}|C{1.4cm}C{1.4cm}C{1.4cm}C{1.4cm}C{1.4cm}C{1.4cm}}
\toprule
\multirow{2}{*}{Layers} & HIS & ChestXray & OCT17 & \multicolumn{2}{c}{BrainMRI} & \multicolumn{2}{c}{LiverCT} & \multicolumn{2}{c}{RESC}\\
\cmidrule(lr){2-10} 
& AC & AC & AC & AC & AS & AC & AS & AC & AS\\
\cmidrule(lr){1-10} 
Layer 1 & 71.19 & 78.80 & 94.82 & 87.47 & 97.13 & 80.04 & \underline{99.67} & 88.03 & 97.15\\
Layer 2 & 80.88 & \textbf{83.56} & \textbf{99.49} & 92.41 & \textbf{97.35} & 80.98 & 99.61 & \underline{95.73} & \underline{98.90}\\
Layer 3 & \underline{82.35} & 58.32 & 96.96 & \textbf{93.01} & 97.14 & 81.10 & 99.59 & 94.18 & \underline{98.90}\\
Layer 4 & 81.43 & 64.58 & 95.36 & \underline{92.86} & 94.43 & \textbf{81.32} & 99.59 & 92.17 & 98.31\\
\cmidrule(lr){1-10} 
Ensemble & \textbf{82.71} & \underline{81.95} & \underline{99.38} & 92.44 & \underline{97.30} & \underline{81.18} & \textbf{99.73} & \textbf{96.18} & \textbf{98.97}\\
\bottomrule
\end{tabular}
}
\end{table*}

\begin{table*}[t]
    \centering
    \caption{Comparisons with state-of-the-art methods on in-domain dataset MVTec AD. The AUCs (in \%) for classification (AC) and segmentation (AS) under the few-shot setting (k=4) are reported.}
    \label{tal:s7_in2}
    \small
    \setlength{\tabcolsep}{2.2pt}{
    \begin{tabular}{C{2.3cm}C{1.4cm}C{1.4cm}C{1.4cm}C{1.4cm}C{1.4cm}C{1.4cm}C{1.4cm}C{1.4cm}}
    \toprule
\multirow{2}{*}{\begin{tabular}{c} 
Category (k=4)
\end{tabular}} & \multicolumn{2}{c}{RegAD} & \multicolumn{2}{c}{ WinCLIP } & \multicolumn{2}{c}{ April-GAN } & \multicolumn{2}{c}{ MVFA } \\
\cmidrule(lr){2-9}  
& AC & AS & AC & AS & AC & AS & AC & AS \\
\cmidrule(lr){1-9} 
bottle & 99.3 & 98.5 & 99.3 & 97.8 & 94.2 & 97.2 & 99.8 & 98.7 \\
cable & 82.9 & 95.5 & 90.9 & 94.9 & 76.7 & 91.8 & 88.0 & 87.3 \\
capsule & 77.3 & 98.3 & 82.3 & 96.2 & 93.5 & 97.5 & 93.9 & 96.0 \\
carpet & 97.9 & 98.9 & 100 & 99.3 & 99.9 & 98.7 & 100 & 99.4 \\
grid & 87.0 & 85.7 & 99.6 & 98.0 & 99.2 & 97.6 & 100 & 96.9 \\
hazelnut & 95.9 & 98.4 & 98.4 & 98.8 & 98.8 & 97.7 & 99.7 & 98.1 \\
leather & 99.9 & 99.0 & 100 & 99.9 & 100 & 99.5 & 99.9 & 99.4 \\
metal nut & 94.3 & 96.5 & 99.5 & 92.9 & 91.0 & 93.1 & 99.4 & 99.3 \\
pill & 74.0 & 97.4 & 92.8 & 97.1 & 84.1 & 95.5 & 95.1 & 96.8 \\
screw & 59.3 & 96.0 & 87.9 & 96.0 & 83.7 & 98.5 & 88.3 & 98.5 \\
tile & 98.2 & 92.6 & 99.9 & 96.6 & 99.1 & 96.0 & 99.7 & 98.7 \\
toothbrush & 91.1 & 98.5 & 96.7 & 98.4 & 93.2 & 98.8 & 95.8 & 98.8 \\
transistor & 85.5 & 93.5 & 85.7 & 88.5 & 84.1 & 83.7 & 84.3 & 80.9 \\
wood & 98.9 & 96.3 & 99.8 & 95.4 & 98.7 & 96.2 & 99.7 & 97.2 \\
zipper & 95.8 & 98.6 & 94.5 & 94.2 & 95.4 & 96.6 & 99.3 & 98.9 \\
\hline average & 89.2 & 96.2 & 95.2 & \textbf{96.3} & 92.8 & 95.9 & \textbf{96.2} & \textbf{96.3} \\
\bottomrule
\end{tabular}}
\end{table*}

\begin{figure*}[h!]
    \centering    
    \includegraphics[width=0.6\textwidth]{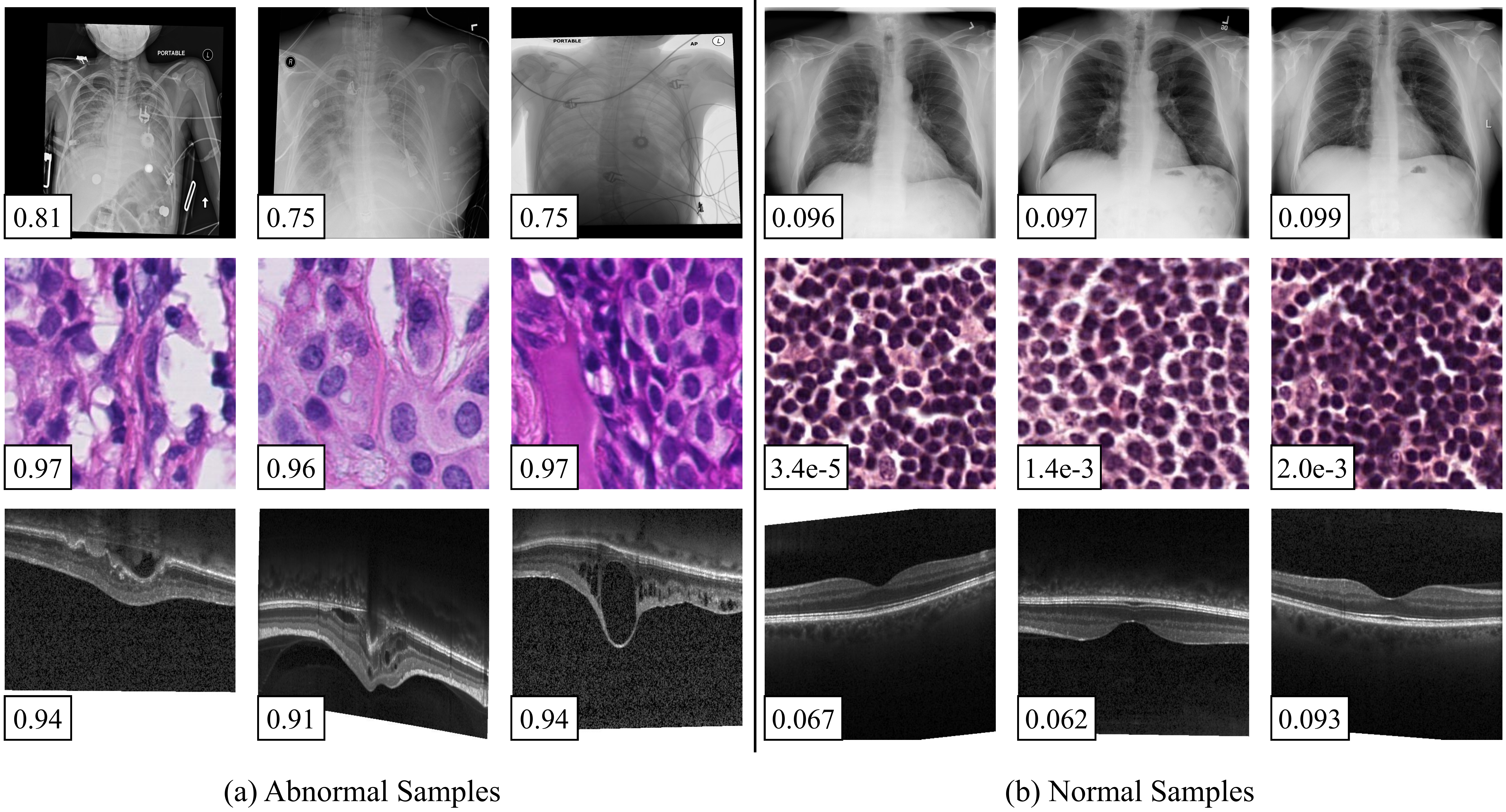}
    \caption{Examples of (a) abnormal samples and (b) normal samples on chest X-ray, histopathology, and retinal OCT. The predicted scores by our method are shown with each sample. The higher the score, the more likely to be an anomaly.}
    \label{fig:visualimg}
\end{figure*}

\begin{figure*}[h!]
    \centering    
    \includegraphics[width=0.75\textwidth]{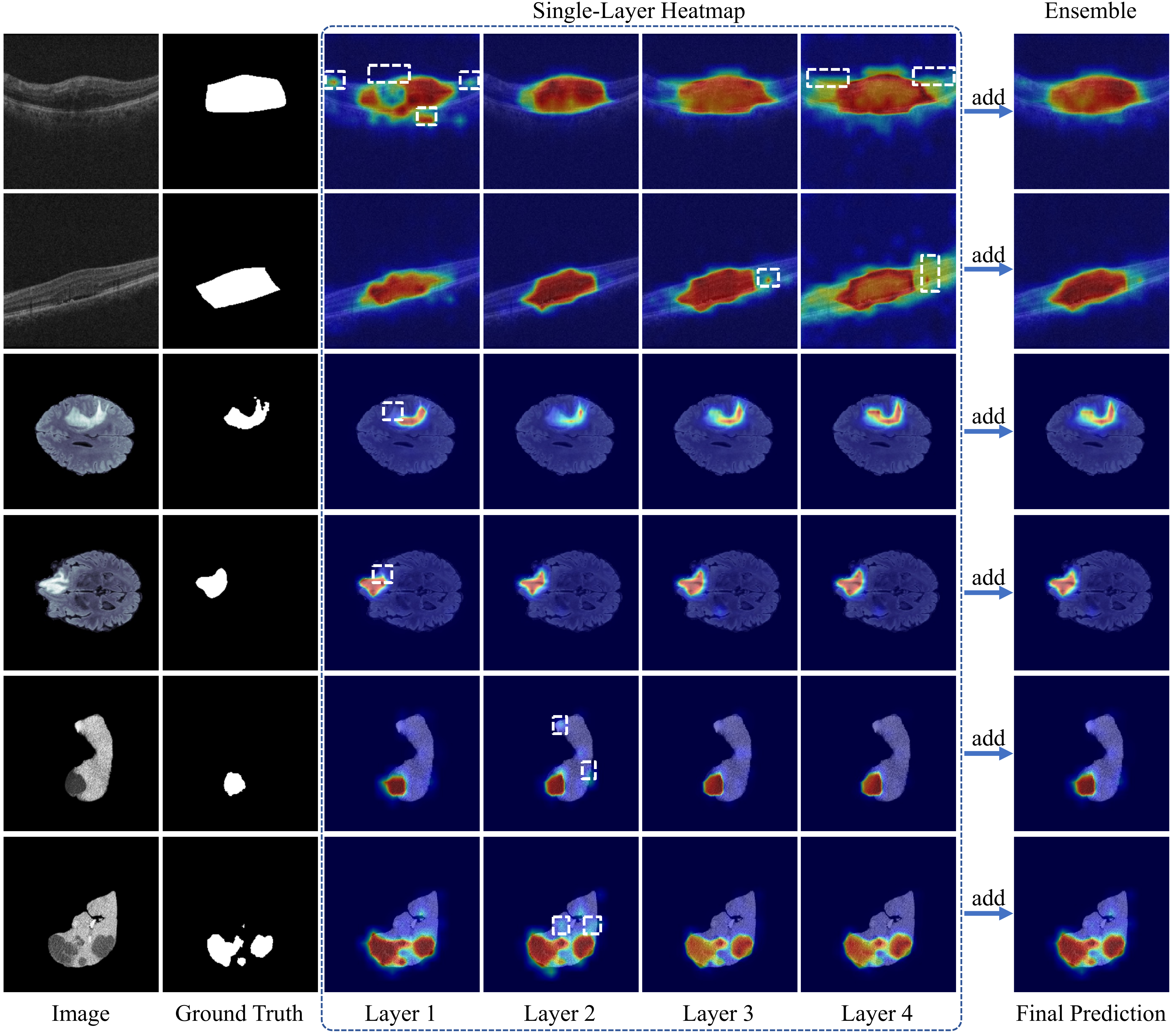}
    \caption{Visualization of anomaly segmentation heatmaps from the four single layers and the multi-layer ensemble results. The white dashed boxes demarcate regions that have been missed or erroneously segmented.}
    \label{fig:multilayer}
\end{figure*}

\begin{figure*}[h!]
    \centering    
    \includegraphics[width=0.75\textwidth]{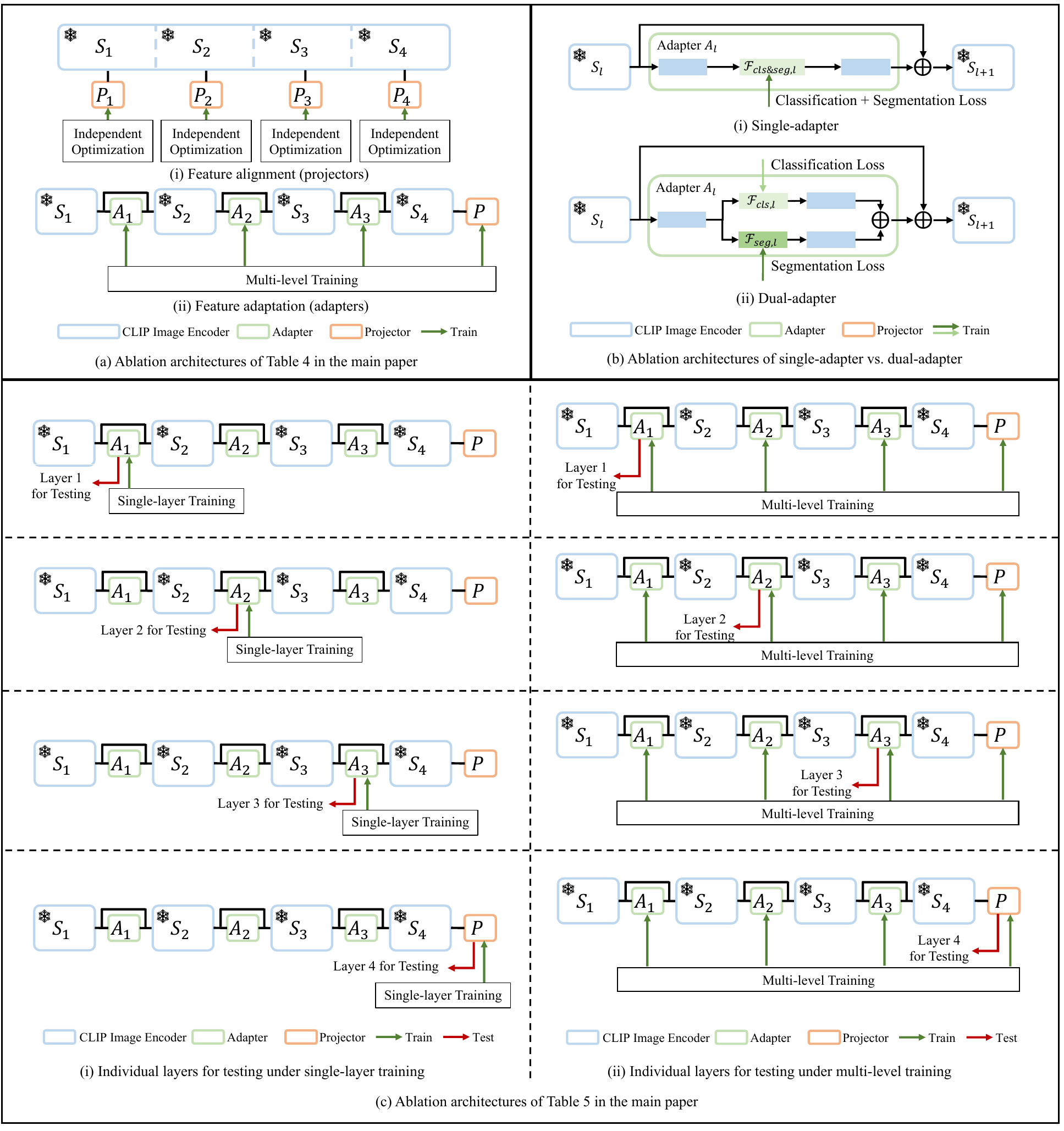}
    \caption{Model structures corresponding to the ablation experimental settings.}
    \label{fig:abla}
\end{figure*}

\noindent\textbf{BrainMRI:} This dataset is built upon the BraTS2021 dataset~\cite{baid2021rsna,bakas2017advancing,menze2014multimodal}, utilizing 3D FLAIR volumes. To account for variations in brain images at different depths, slices within the depth range of 60 to 100 of the 3D FLAIR volumes are selected. Each extracted 2D slice was saved in PNG format and has an image size of $240\times 240$ pixels. The training set encompasses 7,500 normal samples, while the test set comprises 3,715 samples with a balanced ratio of normal to anomaly instances. 

\noindent\textbf{LiverCT:} Derived from two distinct datasets, BTCV~\cite{landman2015miccai} and LiTS~\cite{bilic2023liver}, this dataset is structured to facilitate anomaly detection. The anomaly-free BTCV set, consisting of 50 abdominal 3D CT scans, constitutes the training set, while the test data comprises 131 abdominal 3D CT scans from LiTS. For both datasets, Hounsfield-Unit (HU) of the 3D scans are transformed into grayscale with an abdominal window. The scans are then cropped into 2D axial slices, containing 1,452 2D slices for training and 1,493 2D slices for testing.

\noindent\textbf{Retinal OCT:} The benchmark includes two different OCT AD datasets. The \textbf{RESC} dataset~\cite{hu2019automated} offers pixel-level segmentation labels, delineating regions affected by retinal edema. In contrast, the \textbf{OCT17} dataset~\cite{kermany2018identifying} primarily serves for classification tasks, featuring retinal OCT images categorized into three types of anomalies.

\noindent\textbf{ChestXray:} This dataset comprises lung images, utilizing RSNA~\cite{wang2017chestx} which was originally provided for a lung pneumonia detection task. Abnormal data encompasses cases of ``Lung Opacity'' and cases of ``No Lung Opacity/Not Normal''. The dataset is partitioned into 8,000 normal training images and 17,194 images for testing.

\noindent\textbf{HIS:} Based on Camelyon16~\cite{bejnordi2017diagnostic}, this dataset encompasses 400 whole-slide images (WSIs) of lymph node sections stained with hematoxylin and eosin (H\&E) from breast cancer patients. The training set incorporates 5,088 randomly extracted normal patches from the original training set. For testing, 1,003 normal and 997 abnormal patches from the 115 testing WSIs are utilized.

\section{Text Prompt Formatting}\label{sec:ap_text}
In this study, we adopt a combination of state-level and template-level prompts for generating textual input for the text encoder, as detailed in Figure~\ref{fig:comp_prompt} and following the approach in~\cite{winclip}. The state-level prompts are ingeniously designed by substituting the token [o] with names of human organs such as ``brain'', ``liver'', etc. This substitution strategy allows us to create a varied range of prompts that can categorize images as ``normal'' or ``abnormal'' based on the organ context. We then incorporate these state-level prompts into broader template-level constructs. By replacing the placeholder [c] in a template-level prompt with a corresponding state-level prompt, we formulate prompts that are both comprehensive and contextually rich. This systematic approach enables the creation of detailed, context-specific prompts that accurately distinguish between the normal and abnormal states.

\section{Additional Quantitative Results}\label{sec:ap_results}
\textbf{Results Varied Shot Numbers:} Table~\ref{tal:suppfew} provides a detailed quantitative analysis on the performance of our medical AD approach, benchmarking it against leading few-shot AD methodologies. This analysis is meticulously tabulated, showing our approach's performance specificity across different shot numbers ($K\in\{2,4,8,16\}$). These results lay the groundwork for a line chart featured in the main paper, which visually captures the subtle differences in performance under various conditions.

\noindent \textbf{Ablation Study on Multi-level Features:} 
We carried out an extensive ablation study to evaluate the effectiveness of utilizing single-layer features for each dataset, in line with the average performances across all datasets discussed in the main paper. The outcomes, elucidated in Table~\ref{tal:supplayer_2}, provide a comprehensive understanding of the performance of single-layer features obtained without the implementation of multi-level training. In contrast, Table~\ref{tal:supplayer} presents the results attained through the strategic implementation of multi-level training techniques.

\noindent \textbf{Evaluation on Industrial Anomaly Detection:} For in-domain evaluation, the MVTec AD benchmark~\cite{bergmann2019mvtec}, consisting of 15 industrial defect detection sub-datasets, is considered. MVFA significantly outperforms competing methods, highlighting its superior generalization capabilities. Detailed results for each sub-dataset are included in Table~\ref{tal:s7_in2}.

\section{Additional Qualitative Results}\label{sec:ap_vis}

\noindent \textbf{Anomaly Classification Instances:} Figure~\ref{fig:visualimg} displays results of AC from datasets that only provide anomaly classification labels. These results were obtained using our method in a few-shot setting (K=4). Each instance is accompanied by a predicted score, ranging from zero to one, where higher scores indicate a higher likelihood of an anomaly.

\noindent \textbf{Ensemble of Multi-level Features:} 
Figure~\ref{fig:multilayer} showcases visualizations from different layers used in the anomaly segmentation task. These visualizations include results from datasets such as BrainMRI, LiverCT, and RESC.

\section{Ablation Model Structure}

To effectively convey the nuances of our ablation study in the main paper, we utilized Figure~\ref{fig:abla} to graphically demonstrate the configurations of the models used in our experiments. Specifically, Figure~\ref{fig:abla} (a) visually details the designs of both the adapter and projector as outlined in Table~\ref{tal:abl1} of the main paper, where part (i) illustrates the projector and part (ii) depicts the adapter. In Figure~\ref{fig:abla} (b), we present the configurations for both the single-adapter and dual-adapter models, shown in subfigures (i) and (ii) respectively. Furthermore, Figure~\ref{fig:abla} (c) illustrates the testing pipeline for assessing the impact of training at different levels. Subfigure (i) represents the scenario of single-layer training, while subfigure (ii) demonstrates the approach for multi-level training, corresponding to the discussions and findings presented in Table~\ref{tal:abl3} of the main paper.

\begin{table}[t]
\centering
\caption{Ablation studies of the architecture of dual-adapter against single-adapter in MVFA. The AUCs (in \%) for classification (AC) and segmentation (AS) under the few-shot setting (K=4) are reported, with the best result marked in bold.}
\label{tal:abl2}
\small
\setlength{\tabcolsep}{1.8pt}{
\begin{tabular}{C{2.0cm}|C{1.4cm}C{1.4cm}C{1.4cm}C{1.4cm}}
\toprule
\multirow{2}{*}{Datasets} & \multicolumn{2}{c}{AC} & \multicolumn{2}{c}{AS}\\
& single-adapter & dual-adapter & single-adapter & dual-adapter\\
\cmidrule(lr){1-5}
HIS & 80.80 & \textbf{82.71} & - & - \\
ChestXray & 78.02 & \textbf{81.95} & - & - \\
OCT17 & \textbf{99.87} & 99.38 & - & - \\
BrainMRI & 92.28 & \textbf{92.44} & 96.98 & \textbf{97.30} \\
LiverCT & 81.07 & \textbf{81.18} & 99.42 & \textbf{99.73} \\
RESC & 94.06 & \textbf{96.18} & 98.53 & \textbf{98.97} \\
\cmidrule(lr){1-5}
average & 87.68 & \textbf{88.97} & 98.31 & \textbf{98.67} \\
\bottomrule
\end{tabular}}
\end{table}

\noindent\textbf{Dual-Adapter vs. Single-Adapter.} We compare the performance of the dual-adapter architecture against the single-adapter setup within the few-shot setting. The dual-adapter design, as implemented in our MVFA model, generates two parallel sets of features at each level, catering to both global (classification) and local (segmentation) aspects. The corresponding architectures are shown in Figure~\ref{fig:abla} (b). According to the results in Table~\ref{tal:abl2}, the dual-adapter approach outperforms the single-adapter model on almost all the datasets. We observed an enhancement in the average AUC for AC, improving from 87.68\% to 88.97\%, and for AS, rising from 98.31\% to 98.67\%. This improvement indicates that the dual-adapter architecture is more effective in managing the demands in medical images.

% WARNING: do not forget to delete the supplementary pages from your submission 
% \input{sec/X_suppl}

\end{document}